\documentclass[11pt]{article}

\usepackage[final]{acl}

\usepackage[whole]{bxcjkjatype}

\usepackage{times}
\usepackage{latexsym}

\usepackage[T1]{fontenc}

\usepackage[utf8]{inputenc}

\usepackage{inconsolata}

\usepackage{graphicx}

\usepackage{booktabs}
\usepackage{multirow}

\usepackage{grffile}
\usepackage{graphicx}
\usepackage{verbatim}
\usepackage{comment}
\usepackage{setspace}
\usepackage{adjustbox}
\usepackage{colortbl,array,xcolor}
\usepackage{setspace} 
\usepackage{amssymb}
\usepackage{array}
\usepackage{afterpage} 

\usepackage{amsmath}

\title{Enhancing Persuasive Dialogue Agents \\ by Synthesizing Cross‑Disciplinary Communication Strategies}

\author{ 
    \textbf{Shinnosuke Nozue\textsuperscript{1}\thanks{These two authors contributed equally to this work.}},
    \textbf{Yuto Nakano\textsuperscript{1}\footnotemark[1]},
    \textbf{Yotaro Watanabe\textsuperscript{2}},
    \textbf{Meguru Takasaki\textsuperscript{2}},
    \\
    \textbf{Shoji Moriya\textsuperscript{1}},
    \textbf{Reina Akama\textsuperscript{1,3}},
    \textbf{Jun Suzuki\textsuperscript{1}}
    \\
    \\
    \textsuperscript{1}Tohoku University,
    \textsuperscript{2}PKSHA Technology Inc.,
    \textsuperscript{3}NINJAL,
    \\
    \{nozue.shinnosuke.q5, nakano.yuto.t2, shoji.moriya.q7\}@dc.tohoku.ac.jp,\\
    \{y\_watanabe, meguru\_takasaki\}@pkshatech.com,\\
    \{akama, jun.suzuki\}@tohoku.ac.jp
}

\begin{document}

\maketitle

\begin{abstract}
Current approaches to developing persuasive dialogue agents often rely on a limited set of predefined persuasive strategies that fail to capture the complexity of real-world interactions. 
We applied a cross-disciplinary approach to develop a framework for designing persuasive dialogue agents that draws on proven strategies from social psychology, behavioral economics, and communication theory. 
We validated our proposed framework through experiments on two distinct datasets: the Persuasion for Good dataset, which represents a specific in-domain scenario, and the DailyPersuasion dataset, which encompasses a wide range of scenarios. 
The proposed framework achieved strong results for both datasets and demonstrated notable improvement in the persuasion success rate as well as promising generalizability. 
Notably, the proposed framework also excelled at persuading individuals with initially low intent, which addresses a critical challenge for persuasive dialogue agents.

\end{abstract}

\section{Introduction}
Recent advances in artificial intelligence (AI) research have greatly enhanced the inference and dialogue capabilities of large language models (LLMs)~\citep{brown2020language,openai2023gpt4}, which has increased industry interest in developing dialogue agents capable of persuasion and negotiation to influence human behavior in various social applications~\citep{deng2024towards,deng2025proactive, zhang-etal-2024-strength, zhan-etal-2024-lets, jin-etal-2024-persuading} such as customer support, sales promotion, and healthcare and wellness interventions. 
However, a current research challenge pertains to the comprehensiveness of real-world dialogue strategies.
For instance, recent studies on developing persuasive dialogue agents have often relied only on a subset of annotated persuasive strategies~\citep{song-wang-2024-like, cheng2024cooper, zhang-etal-2024-strength}, such as in the Persuasion for Good (P4G) dataset ~\citep{wang-etal-2019-persuasion} for persuading persuadees to donate to charity. 
This limitation stems from a cross-disciplinary gap, as AI research often overlooks a rich repository of well-established persuasive strategies from practical fields such as sales and marketing, as well as foundational principles from behavioral and social psychology~\citep[e.g.,][]{tversky1981framing,dhar1999effect}. 
Although some research has attempted to expand beyond the predefined annotations in P4G~\citep{jin-etal-2024-persuading}, it still tends to overlook these proven strategies.

To bridge this gap, we take a cross-disciplinary approach, designing a comprehensive strategy that systematically synthesizes a broader range of proven techniques than previously employed~\citep{wang-etal-2019-persuasion,zhang-etal-2024-strength}. 
Our approach systematically synthesizes foundational principles from communication theory, social psychology, and behavioral economics, allowing us to combine powerful techniques to build a more robust and practically grounded framework. 
To evaluate its multifaceted effectiveness, we combined granular intention modeling for analyzing subtle attitude shifts with human assessment of both the dialogue's persuasiveness and its realism.

Through validation on both single and multi-domain datasets, our approach demonstrates notable improvements in persuasion success rates and promising generalizability. 
Our diverse set of strategies is particularly effective on persuadees with low initial intent, often positively shifting their attitudes even in failed dialogues. 
Furthermore, in addition to confirming the method's practical viability through human evaluations, our analysis demonstrates that our proposed extended strategies are actively employed across diverse scenarios and offer new insights into factors influencing optimal strategy selection (such as the persuadee's initial level of intent and conversational context).

\begin{figure*}
    \includegraphics[width=\linewidth]{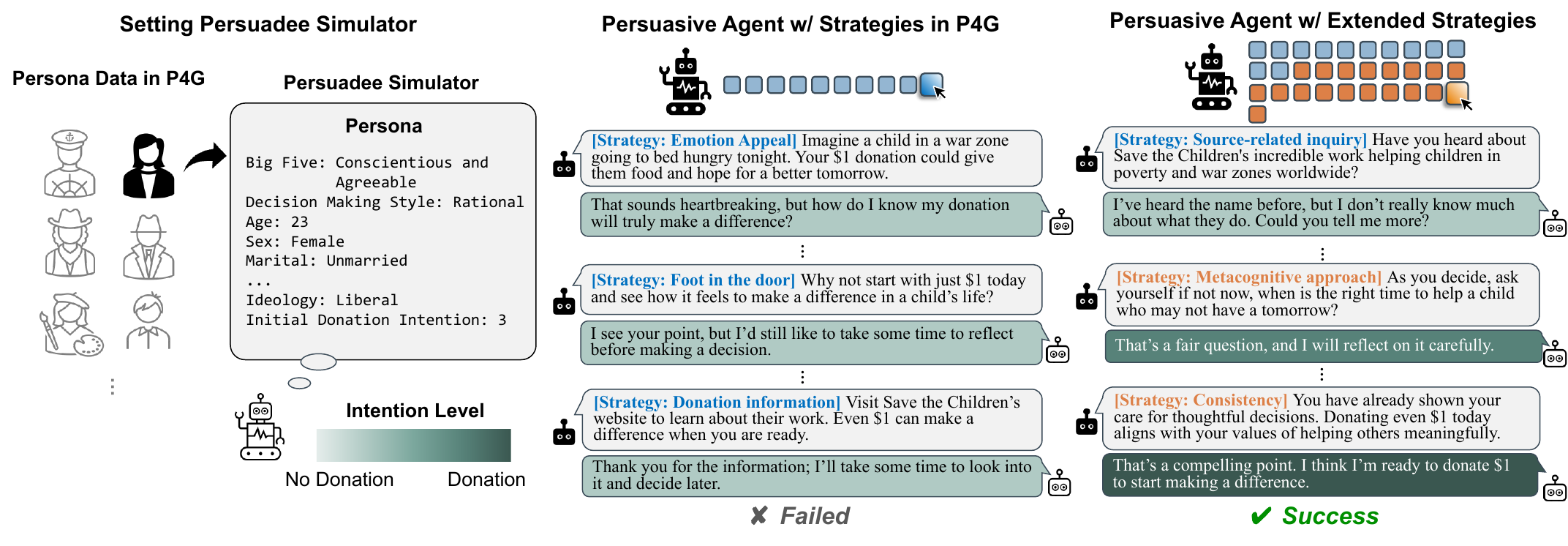}
    \caption{Overview of the proposed framework. To simulate real-world persuasive dialogues, we deploy persuasive dialogue agents equipped with an extended set of empirically validated persuasive strategies (illustrated here for the P4G case).}
    \label{fig:consept}
\end{figure*}

\section{Related Work}
\paragraph{Evaluation Benchmarks and Frameworks.}
Evaluation benchmarks for persuasive and negotiation dialogues have evolved from early datasets with focused scenarios, such as P4G for social good persuasion~\citep{wang-etal-2019-persuasion} and Craigslist Bargain for price negotiation~\citep{he-etal-2018-decoupling}. 
More recent efforts have expanded this scope, with comprehensive interactive benchmarks, such as SOTOPIA~\citep{zhou2024sotopia}, for assessing broad social intelligence through diverse tasks, and large-scale LLM-generated datasets, like DailyPersuasion~\citep{jin-etal-2024-persuading}, that cover a multitude of domains. 
The rise of such generated data has also spurred the development of evaluation benchmarks for ensuring data faithfulness~\citep{zhang2025doubleblind} and assessing latent cognitive states~\citep{yu2025persuasivetom, bozdag2025persuade}.

\paragraph{Approaches to Persuasive Dialogue Agents.}
Early work on persuasive dialogue agents addressed a range of sub-problems, from classifying the resistance strategies of the persuadee~\citep{tian-etal-2020-understanding} to modeling the strategic choices of the persuader with classification~\citep[e.g.,][]{saha-etal-2021-performance, mishra-etal-2022-pepds} or rules~\citep[e.g.,][]{chaves2021effects}. 
Recent studies on developing LLM-based persuasive dialogue agents have diverged into two approaches.  
The first approach involves complex learning-based adaptation, which includes methods that leverage reinforcement learning (e.g., TRIP~\citep{zhang-etal-2024-strength}, PPDPP~\citep{deng2024plug}, and DPDP~\citep{he-etal-2024-planning}) and extends to sophisticated frameworks that incorporate persuadee modeling~\citep{he2025simulating}, latent policy discovery~\citep{he2025simulationfree}, or causal inference~\citep{zeng2025causal}. 
While powerful, these approaches introduce substantial architectural complexity and training costs.
In contrast, the second approach uses efficient, training-free prompting. 
However, a core limitation of previous methods has been their reliance on either a narrow set of strategies often confined to a single academic domain, as seen in many P4G-based systems~\citep{cheng2024cooper, song-wang-2024-like}, or fundamental strategies that overlook the nuanced techniques proven in practice in fields like social psychology and behavioral economics~\citep{jin-etal-2024-persuading}. This highlights the need for a more knowledge-driven approach to achieve greater persuasive nuance and avoids the substantial architectural complexity and training costs mentioned earlier.

\section{Methodology}
\label{sec:procot-rich}
This section details the methodology for evaluating persuasive dialogue agents, focusing on
an extended set of persuasive strategies that we developed based on behavioral and social psychology. 
To operationalize and test these novel strategies, we utilize an LLM to interpret the dialogue context and dynamically generate persuasive utterances.
The entire framework, illustrated in Figure~\ref{fig:consept}, enables a robust and comprehensive examination of persuasive efficacy using our extended strategy set across multiple dialogue datasets.

\subsection{Cross-Disciplinary Persuasive Strategies}
To build a practically effective agent, we moved beyond the predefined strategies of datasets like P4G~\citep{wang-etal-2019-persuasion} and constructed a cross-disciplinary framework. Our approach systematically synthesizes foundational principles from communication theory, social psychology, and behavioral economics. 
While P4G includes 10 persuasive strategies\footnote{``Logical appeal,'' ``Emotion appeal,'' ``Credibility appeal,'' ``Foot in the door,'' ``Self-modeling,'' ``Personal story,'' ``Donation information,'' ``Source-related inquiry,'' ``Task-related inquiry'' and ``Personal-related inquiry.''}, it excludes many well-established strategies from these other disciplines.

Our framework first draws on communication theory and social psychology to model the core persuasive process on the Elaboration Likelihood Model (ELM)~\citep{petty1986communication}, which differentiates between the central route (e.g., ``Logical appeal'') for durable attitude change and the peripheral route (e.g., ``Emotion appeal'') for a quicker influence. 
The Heuristic-Systematic Model (HSM)~\citep{chaiken1980heuristic} is also utilized to support our use of multiple strategies within a single dialogue. 
To promote specific actions, while P4G includes ``Foot in the door''~\citep{freedman1966compliance}, we incorporate other powerful techniques such as ``Door in the face''~\citep{cialdini1975reciprocal} and ``Reciprocity''~\citep{cialdini2001science}.

Second, our framework incorporates insights from behavioral economics on how the presentation of information influences decision-making. 
These strategies include ``Framing''~\citep{tversky1981framing}, which changes how the information is presented to alter a decision, and the ``Principle of Scarcity''~\citep{Cialdini1984, cialdini2001science}, which prompts action by highlighting potential missed opportunities. 
Another effective strategy is ``Time pressure,'' which imposes a temporal constraint~\citep{kruglanski1983freezing}, particularly in situations with significant option conflicts~\citep{dhar1999effect}.

Thus, our framework broadly classifies persuasive strategies into the following categories: inquiries to persuadees, basic persuasion techniques, trust and credibility building, promotion of specific actions, information presentation techniques, personalization, and follow-up strategies. 
Based on these categories, the framework can be used to navigate key stages of the persuasion process from initial engagement to sustained behavioral changes.

\subsection{Relation with the P4G Strategy Set}
Our cross-disciplinary framework expands the 10 strategies of P4G into a more comprehensive set of 31 strategies, organized into seven distinct categories. 
While most of the P4G strategies are included directly, a few are mapped to new labels that better fit the defined categories. 
For example, the original self-modeling strategy was renamed as ``personal demonstration.''
A detailed mapping and explanation of how our framework expands upon the original P4G strategies is provided in Table~\ref{tab:strategy_list} in the Appendix.

\subsection{Strategy Implementation Method}
To efficiently implement our proposed framework, we employed the proactive Chain-of-Thought prompting (ProCoT) scheme~\citep{deng-etal-2023-prompting}. 
This scheme employs a Chain-of-Thought (CoT) process~\citep{10.5555/3600270.3602070}, instructing the LLM first to interpret the dialogue history, then infer a suitable strategy, and finally generate a response. 
The prompt provided to the model contains all the necessary components for this task: a task overview, output constraints, a list of persuasive strategies, and the whole dialogue history. 
This enables the model to generate utterances based on its inferred strategy autonomously.
While we opted for a prompting approach because of its efficiency, the set of strategies within our proposed framework is method-independent. It can be integrated into other approaches, including those based on reinforcement learning.

\setlength{\tabcolsep}{3.1pt} 

\begin{table*}[ht]
    \footnotesize
    \centering
    \begin{tabular}{lcccc ccccc ccccc}
        \toprule
                \bf{Agent Pattern} & \bf{SR} $^\uparrow$ & \bf{AT} $^\downarrow$ & \bf{AT-SD} $^\downarrow$ & \bf{AII} $^\uparrow$ & \bf{SR1} & \bf{SR2} & \bf{SR3} & \bf{SR4} & \bf{SR5} & \bf{AII1}  & \bf{AII2} & \bf{AII3}  & \bf{AII4} & \bf{AII5} \\
        \cmidrule(l){1-1} \cmidrule(l){2-5} \cmidrule(l){6-10} \cmidrule(l){11-15}
        Simple & 0.553 & 7.383 & 5.271 & 0.261 & 0.959 & 0.946 & 0.446 & 0.245 & 0.081 & -1.000 & 0.000 & 0.387 & 0.375 & 0.193 \\
        ProCoT-p4g & 0.760 & 6.310 & 5.145 & 0.417 & 0.986 & \bf{1.000} & 0.804 & 0.642 & 0.339 & -1.000 & - & 0.273 & 0.474 & \bf{0.463} \\
        ProCoT-p4g-desc & 0.707 & 6.990 & 5.741 & 0.318 & \bf{1.000} & 0.946 & 0.821 & 0.472 & 0.242 & - & 0.000 & \bf{0.500} & 0.536 & 0.170 \\
        \cmidrule(l){1-1} \cmidrule(l){2-5} \cmidrule(l){6-10} \cmidrule(l){11-15}        ProCoT-rich & 0.830 & \bf{5.707} & \bf{4.827} & 0.216 & \bf{1.000} & \bf{1.000} & \bf{0.893} & 0.774 & 0.468 & - & - & 0.167 & 0.417 & 0.152 \\
        ProCoT-rich-desc & \bf{0.833} & 6.057 & 5.268 & \bf{0.480} & \bf{1.000} & 0.946 & \bf{0.893} & \bf{0.811} & \bf{0.500} & - & 0.000 & 0.333 & \bf{1.100} & 0.355 \\
        \bottomrule
    \end{tabular}
    \caption{Performance by persuasive agent patterns. SR: success rate of persuasion, AT: average turns until success, AT-SD: AT in successful dialogues only, AII: average intention improvement in failed dialogues.}
    \label{tab:result}
\end{table*}

\setlength{\tabcolsep}{6pt} 

\section{Experimental Settings}
\subsection{Persuasive Dialogue Agents}
We developed two agents to serve as baselines: \textbf{\textit{Simple}} was provided only with the task overview and dialogue history, and \textbf{\textit{ProCoT-p4g}} leveraged strategy annotations from P4G.  
The proposed framework detailed in Section \ref{sec:procot-rich} was then used to develop \textbf{\textit{ProCoT-rich}}. 
To assess the impact of strategy interpretability in the prompts, we created two additional agents that included strategy descriptions in their prompts: \textbf{\textit{ProCoT-p4g-desc}} and \textbf{\textit{ProCoT-rich-desc}}. 
The strategy descriptions for \textit{ProCoT-p4g-desc} were based on previous work~\citep{wang-etal-2019-persuasion}. 
In contrast, \textit{ProCoT-p4g} and \textit{ProCoT-rich} (without ``-desc'' agents) were provided only with strategy labels. 
We used OpenAI's \texttt{gpt-4o-2024-11-20} for all the persuasive dialogue agents.


\subsection{Evaluation for P4G Dataset}

To evaluate the in-domain performance of the persuasive dialogue agents, we tested them on 300 samples from the P4G dataset. 
We utilized both automatic and human evaluation methods to assess their performance.

\paragraph{Automatic Evaluation.}
\label{sec:persuadee_simulator}
For the automatic evaluation, we developed an LLM-based approach to generate simulated persuadee prompted with task overviews, personas\footnote{See Appendix \ref{sec:persona-setting} for more details}, and dialogue history~\citep{zhang-etal-2024-strength}. 
Initially, instructing the persuadee simulators to resist persuasion led to biased and repetitive utterances, so we stopped enforcing predefined resistance strategies.
We used OpenAI's \texttt{gpt-4o-2024-11-20} as our persuadee simulator.

We defined the success of persuasion more strictly than previous work. 
While \citet{zhang-etal-2024-strength} counted any positive response as a success and \citet{yu-etal-2023-prompt} used a cumulative scoring system, we required an explicit intention to donate. 
Using the five-level intention scale from \citet{yu-etal-2023-prompt}, we employed an LLM to evaluate the persuadee's intention to donate at each turn, considering both their persona and dialogue history. 
Each turn was evaluated 10 times, and a dialogue was considered successful if the majority indicated level-1 (Donate).
The evaluation continued for up to 10 turns and ended upon success.
The initial intention to donate was also set on a five-level scale\footnote{level-1 (Very Keen), level-2 (Keen), level-3 (Undecided), level-4 (Initially Reluctant), and level-5 (Explicit Non-donor)}.
We used \texttt{gpt-4o-mini-2024-07-18} as the evaluation model\footnote{Following \citet{deng-etal-2023-prompting}, we set the temperature to 0 to minimize output variability.\label{fot:temp-0}}.

We evaluated the persuasive dialogue agents using several metrics. 
Our primary measure was the \textbf{success rate (SR)}, the proportion of dialogues that successfully achieved the given task. 
We also assessed SR according to the initial level for the intention to donate.
For example, SR1 indicates the success rate for persuadees whose initial donation intention was level-1 (Very Keen).
To assess efficiency, we also calculated the \textbf{average turns (AT)}, which is the average number of turns across all dialogues. 
To mitigate bias from interactions that failed early, we also measured \textbf{AT in successful dialogues (AT-SD)}, which we defined as the average number of turns in successful dialogues only.
Finally, we introduced the \textbf{Average Intention Improvement (AII)} to measure changes in the persuadee's intention to donate during unsuccessful dialogues:
\begin{gather}
    \textbf{AII} = {\sum}_{i=1}^N \frac{I^{\text{initial}}_{i} - I^{\text{final}}_{i}}{N}
\end{gather}
where $I^{\text{initial}}_{i}$ and $I^{\text{final}}_{i}$ represent the intentions to donate before and after the $i$-th unsuccessful dialogue, respectively, and $N$ denotes the total number of unsuccessful dialogues.

\paragraph{Human Evaluation.}
For the human evaluation, we asked annotators to assess dialogues in two dimensions: \textbf{persuasiveness} and \textbf{realism}.
Regarding persuasiveness, the annotators selected their preferred response from dialogue pairs generated by \textit{ProCoT-p4g} and \textit{ProCoT-rich-desc}. 
Regarding realism, the annotators were asked to rate how closely dialogues from \textit{ProCoT-rich-desc} resembled natural human conversation on a five-point Likert scale, where 1 represented ``very unrealistic'' and 5 represented ``very realistic.''
Both evaluations were conducted independently by two annotators, who each assessed 300 dialogues each.

\subsection{Evaluation for DailyPersuasion}
We evaluated the generalized persuasive performance of the agents by testing them on 1000 samples from the DailyPersuasion dataset. 
For this evaluation, we compared \textit{ProCoT-rich-desc} against \textit{Simple} and \textit{ProCoT-p4g} as the baselines.

The DailyPersuasion dataset encompasses 13,000 scenarios across 35 distinct domains, with each scenario containing an average of six dialogues. 
Each dialogue begins with a statement from the persuasive dialogue agent and includes alternating exchanges between the agent and the persuadee simulator. 
On average, dialogues include five turns, with a maximum of 16 turns per dialogue.
To achieve a diverse evaluation, 1,000 scenarios were randomly selected at the scenario level. 
From each selected scenario, one dialogue was randomly chosen. 
Finally, for each dialogue, the number of previous turns included as the dialogue history was randomly determined, ranging from zero to one less than the total number of turns in the dialogue\footnote{Appendix~\ref{sec:daily-persuasion-data} details dataset statistics.}.

First, the persuasive dialogue agent received the task scenario, goal, and dialogue history up to the previous turn.
It then outputs a persuasive utterance in response to the last utterance of the persuadee in the dialogue history.
Next, we provided the evaluation instructions, scenario, goal, dialogue history, and two anonymized agent responses, arranged in a random order, to a judgment model, using \texttt{o3-2025-04-16}\footnote{The reasoning effort was set to ``high.''}.
The judgment model could designate one response as the winning response or classify both responses as either acceptable (Comparable-Good) or unacceptable (Comparable-Bad). 
The comparable-good and comparable-bad labels were treated as ties.
To avoid position bias, each pair of responses was judged twice in reverse order, and inconsistent judgments were marked as a tie.
The dialogues covered 35 diverse domains, with the target turns ranging from the first to the eighth turn.

We used the win rate as the evaluation metric, which we defined as the percentage of times that the response of \textit{ProCoT-rich-desc} was judged superior to those of the baselines.

\section{Results and Discussion}
\subsection{Experimental Results}
Table~\ref{tab:result} presents the automatic evaluation results for the P4G dataset\footnote{The English version is presented in Appendix~\ref{sec:en-result}.}.
\textit{ProCoT-rich-desc} achieved the highest SR, which indicates its strong persuasive ability. 
Across different initial intention levels, all agents achieved near-perfect success when the persuadees had a high intention to donate (SR1, SR2). 
In contrast, \textit{ProCoT-rich-desc} was more effective than the baselines when the initial intention levels were lower (SR4, SR5), which suggests that it is particularly effective on persuadees with a very low willingness to donate.
\textit{ProCoT-rich-desc} also achieved the highest AII and reached an AII4 of 1.100, which implies that it can increase the intention to donate among persuadees who initially had a low but non-zero interest.
Regarding AT and AT-SD, \textit{ProCoT-rich} exhibited the best performance, which indicated that its dialogues were more efficient.
Although \textit{ProCoT-rich-desc} had a shorter AT than almost all agents, its longer AT-SD suggests a relatively low efficiency during successful dialogues. 

\begin{table}[t]
    \footnotesize
    \centering
    \begin{tabular}{lccc}
        \toprule
        \textbf{Agent Variations} & \textbf{Win (\%)} & \textbf{Tie (\%)} & \textbf{Lose (\%)} \\
        \midrule
        vs. ProCoT-p4g & \textbf{72.5} & 8.8 & 18.7 \\
        \bottomrule
    \end{tabular}
    \caption{Human-evaluated persuasiveness Win-Rate of \textit{ProCoT-rich-desc} over \textit{ProCoT-p4g} on the P4G dataset.}
    \label{tab:result-p4g-humaneval-persuasiveness}
\end{table}

Table~\ref{tab:result-p4g-humaneval-persuasiveness} presents the human evaluation results for persuasiveness on the P4G dataset.
\textit{ProCoT-rich-desc} achieved a win rate of 72.5\%, which indicated that the annotators had a strong preference for it over \textit{ProCoT-p4g}.
Inter-annotator agreement was measured using Cohen's kappa, which yielded a value of 0.375, corresponding to a fair level of consistency.
Regarding realism, the average rating across 600 evaluations was 3.73, which suggests that the dialogues were generally perceived as natural and plausible human conversations.
Qualitative feedback from the annotators revealed that lower ratings often stemmed from the agent imposing suggestions without sufficient empathy or consideration for the persuadee's perspective.
Additionally, the annotators commented on the unnaturalness caused by the repetitive use of similar expressions throughout the dialogue.

\begin{table}[t]
    \footnotesize
    \centering
    \begin{tabular}{lccc}
        \toprule
        \textbf{Agent Variations} & \textbf{Win (\%)} & \textbf{Tie (\%)} & \textbf{Lose (\%)} \\
        \midrule
        vs. Simple & \textbf{54.4} & 30.4 & 15.2 \\
        vs. ProCoT-p4g & \textbf{35.1} & 39.2 & 25.7 \\
        \bottomrule
    \end{tabular}
    \caption{Automatic Evaluation: Win-Rate of \textit{ProCoT-rich-desc} against \textit{Simple} and \textit{ProCoT-p4g} on the DailyPersuasion dataset.} 
    \label{tab:result-daily-persuasion}
\end{table}
Table~\ref{tab:result-daily-persuasion} presents the evaluation results for the DailyPersuasion dataset. 
\textit{ProCoT-rich-desc} outperformed both \textit{Simple} and \textit{ProCoT-p4g}, demonstrating its robustness in generating persuasive responses across diverse domains.

\begin{table}[t]
    \small
    \centering
    \tabcolsep 1pt
    \begin{tabular}{l l r r r}
        \toprule
         & & \multicolumn{2}{c}{\textbf{Number of Tokens}} & \textbf{Response} \\
         & &\textbf{ Input} & \textbf{Output} & \textbf{Time (s)} \\
        \midrule
        \multirow{3}{*}{
        \begin{tabular}{l}
            P4G
        \end{tabular}
        } & Simple & $340.35$ & $30.27$ & $0.84$ \\
          & ProCoT-p4g & $505.26$ & $156.58$ & $2.73$ \\
          & ProCoT-rich-desc & $1272.89$ & $152.55$ & $1.94$ \\
        \midrule
        \multirow{3}{*}{
        \begin{tabular}{l}
            Daily\\Persuasion
        \end{tabular}
        } & Simple & $282.35$ & $25.15$ & $0.87$ \\
          & ProCoT-p4g & $408.35$ & $109.75$ & $1.64$ \\
          & ProCoT-rich-desc & $1208.35$ & $116.87$ & $2.51$ \\
        \bottomrule
    \end{tabular}
    \caption{Average token counts and response times per turn in P4G and DailyPersuasion. The averages were computed from 30 P4G dialogues (6-7 turns each) and 100 single-turn DailyPersuasion dialogues.}
    \label{tab:time_p4g_dp}
\end{table}
Table~\ref{tab:time_p4g_dp} shows the average response times per turn in the P4G and DailyPersuasion datasets.
In each dataset, 10\% of the total instances were sampled to calculate the average token count and average response time per turn.
The \textit{Simple} and \textit{ProCoT-p4g} settings served as baselines, while our proposed \textit{ProCoT-rich-desc} configuration was evaluated for comparison.
Agents with longer input and output lengths, such as \textit{ProCoT-p4g} and \textit{ProCoT-rich-desc}, tend to have longer response times.
The results from P4G further suggest that response time may be more affected by output length than by input length.
However, these results should be interpreted as reference values, as they can be influenced by factors such as API congestion and network conditions.

The increase in complexity also correlated with higher computational costs. 
The average costs per turn were \$0.001, \$0.003, and \$0.005 for \textit{Simple}, \textit{ProCoT-p4g}, and \textit{ProCoT-rich-desc} in P4G, and \$0.001, \$0.002, and \$0.004, respectively, in DailyPersuasion.
This highlights the trade-off between response quality and operational cost.

For fine-grained analysis, we focused on the P4G dataset as its single-domain nature provides a controlled setting ideal for investigating details such as strategy usage and the effects of the persuadee's intention level. 
A detailed analysis of the DailyPersuasion dataset is presented in Appendix~\ref{sec:daily-persuasion-analysis}.

\paragraph{Usage of Persuasive Strategies.}
\begin{table}[t]
    \centering
    \small
    \tabcolsep 3pt
    \begin{tabular}{lcccc}
        \toprule
        \bf{Persuasive Strategy} & \bf{PC-p} & \bf{PC-pd} & \bf{PC-r} & \bf{PC-rd} \\
        \midrule
        Emotion appeal & 0.488 & 0.281 & 0.282 & 0.247 \\ 
        Foot in the door & 0.221 & 0.144 & 0.241 & 0.241 \\ 
        Donation information & 0.131 & 0.128 & - & - \\ 
        Credibility appeal & 0.063 & 0.189 & 0.038 & 0.058 \\ 
        Source-related inquiry & 0.006 & 0.151 & 0.000 & 0.168 \\ 
        Detailed information & - & - & 0.099 & 0.040 \\ 
        Logical appeal & 0.039 & 0.048 & 0.031 & 0.127 \\ 
        Personal-related inquiry & 0.001 & 0.004 & 0.171 & 0.000 \\ 
        Social proof & - & - & 0.054 & 0.004 \\ 
        Time pressure & - & - & 0.042 & 0.016 \\ 
        Self-modeling & 0.029 & 0.023 & - & - \\ 
        Feedback and thanks & - & - & 0.017 & 0.023 \\ 
        \midrule
        Entropy (w/o unused str.) & 2.167 & 2.738 & 2.871 & 2.973 \\
        Entropy (all) & 2.157 & 2.721 & 0.194 & 0.181 \\
        \bottomrule
    \end{tabular}
    \caption{Distributions of strategies used by each agent. The top part of the table shows the proportion of each strategy used among all utterances made by the agent.
    \textbf{PC-p}, \textbf{PC-pd}, \textbf{PC-r}, and \textbf{PC-rd} represent the shortened notations for \textbf{\textit{ProCoT-p4g}}, \textbf{\textit{ProCoT-p4g-desc}}, \textbf{\textit{ProCoT-rich}}, and \textbf{\textit{ProCoT-rich-desc}}, respectively.}
    \label{tab:agent_strategy_analisis}
\end{table}
Table~\ref{tab:agent_strategy_analisis} presents the proportion of strategies employed by each persuasive dialogue agent.
Because \textit{ProCoT-rich-desc} incorporated all 10 of the original P4G strategies (albeit with some relabeling\footnote{Table~\ref{tab:strategy_list} in Appendix presents the complete mapping between our terminology and the original labels of P4G.}), a blank entry for this agent indicates that an available strategy was not used. 
To quantify the bias in each agent's strategy usage, we calculated the entropy from the distribution of strategies.
When all available strategies were considered, \textit{ProCoT-rich} and \textit{ProCoT-rich-desc} exhibited lower entropy than \textit{ProCoT-p4g}, which indicates less diverse strategy use.
However, when only the strategies employed were considered, both \textit{ProCoT-rich} and \textit{ProCoT-rich-desc} displayed relatively high entropy.
Among these, \textit{ProCoT-rich-desc} demonstrated higher entropy than \textit{ProCoT-rich} and performed better at increasing the persuadee's intention to donate as measured by SR and AII. 
Conversely, while \textit{ProCoT-p4g-desc} demonstrated a higher entropy than \textit{ProCoT-p4g}, its overall performance remained lower. 
These results suggest that greater diversity in strategy selection influenced by the prompt's interpretability does not necessarily improve persuasive efficacy.

\paragraph{Effective Strategies by Donation Intention Level.}
\begin{figure}[t]
    \centering
    \includegraphics[width=\linewidth]{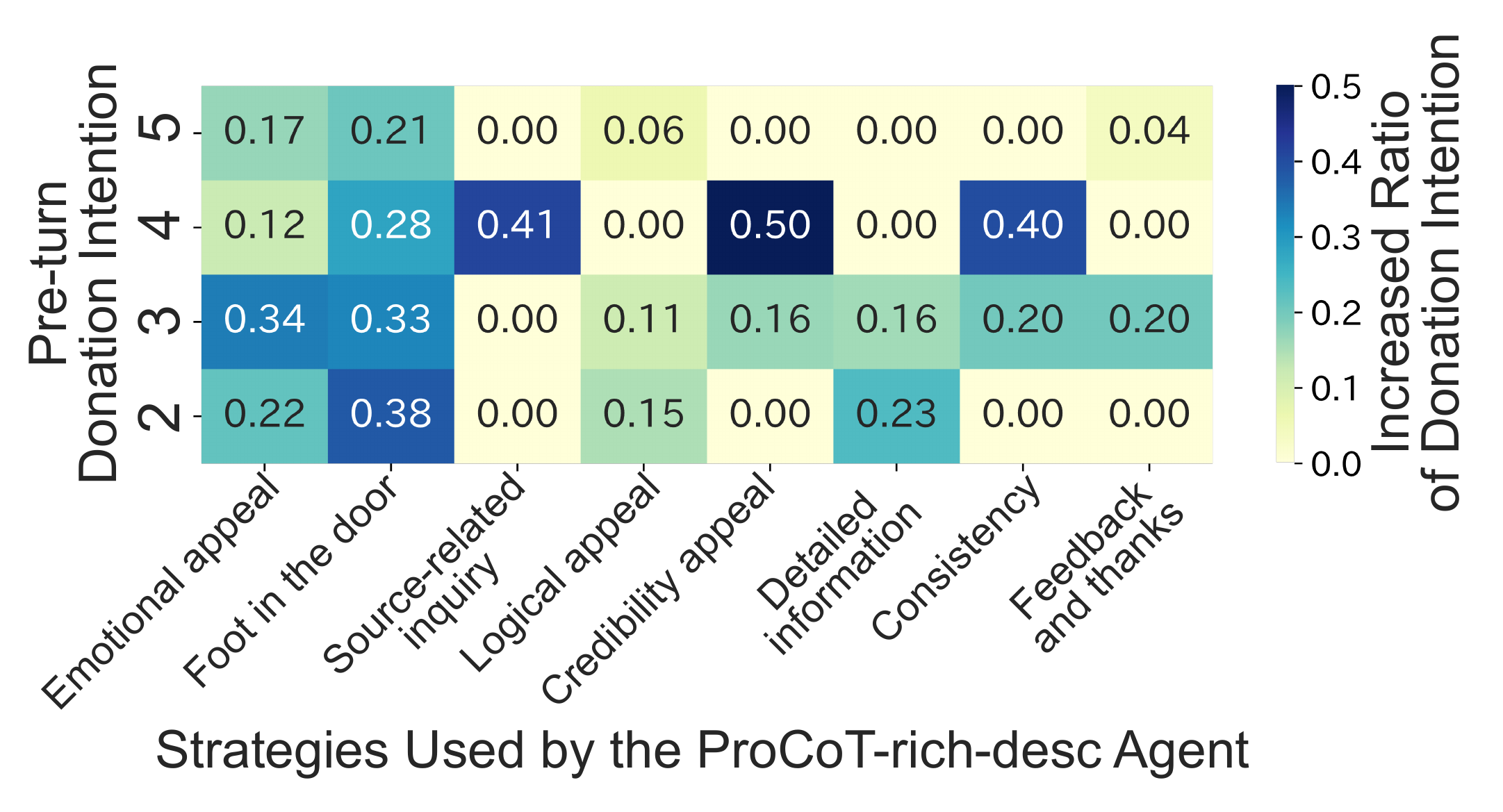}
    \caption{Proportion of cases in which each strategy increased the intention of persuadees to donate at each donation intention level.}
    \label{fig:heat_map}
\end{figure}

To explore how persuasive strategies influenced the intention to donate, we analyzed strategies employed by \textit{ProCoT-rich-desc} at least 40 times, as visualized in Figure~\ref{fig:heat_map}.
Among persuadees with a level-4 (Negative Reaction), a narrow set of strategies proved effective, especially ``Credibility appeal,'' ``Source-related inquiry,'' and ``Consistency.'' 
Among persuadees with level-5 (No Donation), ``Foot in the door'' and ``Emotion appeal'' provided limited but measurable improvements.
``Foot in the door'' was also effective on persuadees with a high intention to donate. 
However, requesting a small donation from those already inclined to donate may be a missed opportunity to secure a larger donation.

\paragraph{Impact of Granular Intention Level.}
Granular segmentation of the initial intention allowed for more precise targeting of persuadee subgroups, such as those with negative intention but potential for change. 
This granularity has notable implications for practical applications.
For example, while 23.5\% of persuadees classified as negative by conventional approaches - specifically level-4 (Negative Reaction) and level-5 (No Donation) - exhibited no change in intention, narrowing the scope to level-4 (Negative Reaction) reduced this to just 3.7\%. 
This refined approach enables the potential for Return On Investment (ROI)-optimized persuasion. 
Developing persuasive strategies tailored for level-5 (No Donation) remains a key research challenge\footnote{Further details are provided in Table~\ref{tab:intention-shifts} in Appendix~\ref{seq:appendix_analysis_of_shifting}.}.

\section{Conclusion and Future Work}
\label{sec:conclusion_and_future_work}
This paper introduced a framework of cross-disciplinary persuasive strategies, synthesizing insights from diverse fields, and evaluated their effectiveness using automatic evaluation with realistic persuadee simulators and manual evaluation by humans.
Our results confirmed the effectiveness of these extended persuasive strategies, particularly for persuadees with low initial donation intentions.
Additionally, the findings highlight that the optimal persuasive strategy varies based on the strength of the initial donation intention. 

Several remaining challenges were identified. 
First, it became clear that strategies were not being adapted appropriately to the context.
Future work should explore methods for selecting context-appropriate strategies, such as integrating a strategy estimation agent into the multi-agent framework.
Second, a more comprehensive evaluation is needed. 
The current evaluation did not consider the amount of the donation, despite this being an important factor. 
In cases where a strategy such as foot-in-the-door secures at least a small donation, integrating additional strategies like door-in-the-face may yield higher donation amounts. However, such combinations were not observed in this study. Thus, future work should explore the potential impact of combining multiple strategies to maximize donation outcomes, and future evaluations should include the donation amount. Moreover, because our current evaluation focused solely on shifts in intention, future works should introduce comprehensive metrics that account for both. 
Finally, a crucial next step is to validate the practical utility of our framework with human participants in real-world business domains. 
We plan to explore its applicability in real-world business domains, such as providing advanced support for operators in call centers and assisting with dialogue in the sales domain. 
These real-life case studies will be essential to bridge the gap between our simulation-based findings and real-world applications.

\section*{Limitations}
While our experimental results demonstrated the effectiveness of our proposed framework through simulated dialogues, they come with several limitations.

\paragraph{Reliance on Simulation-Based Evaluation and Ecological Validity.}
The evaluations in this study are primarily based on dialogue simulations with LLMs on the P4G and DailyPersuasion datasets. 
While this approach allows for a large-scale and reproducible evaluation, it does not fully replicate the complexity and unpredictability of interactions with real humans. 
A potential bias may arise from using the same family of models (gpt-4o-2024-11-20) for both the persuasive dialogue agent and the persuadee simulator. 
As noted in Section~\ref{sec:conclusion_and_future_work}, experiments with actual human participants are essential to validate the true effectiveness of our proposed framework.

\paragraph{Limitations of Evaluation Metrics and Methods.}
We primarily measured success by using metrics such as SR and AII. 
However, these metrics do not capture the qualitative aspects of persuasion. 
For instance, in the context of the P4G dataset, the donation amount was not considered, and donations of \$1 and \$100 were treated equally as successes.
Furthermore, the evaluation of the DailyPersuasion dataset relies on an LLM-as-judge (Win-Rate), which may not fully reflect the nuances of human perception of persuasiveness and realism. 
The ``fair'' level of inter-annotator agreement (Cohen's kappa of 0.375) for the human evaluation of the P4G dataset also suggests that assessing persuasiveness is an inherently subjective and challenging task.

\paragraph{Comprehensiveness and Combination of the Strategy Set.}
While a key contribution of this work is extending the set of persuasive strategies based on cross-disciplinary insights, it is difficult to claim that the new list is fully comprehensive. 
Moreover, strategies are evaluated in isolation within a single turn. 
In real-world situations, strategies are often skillfully combined or sequenced. 
However, the effectiveness of such strategy combinations (e.g., using ``Door in the face'' followed by a concession) was not explored and remains an area for future work.

\paragraph{Scope of Generalizability.}
The fine-grained analysis on the effectiveness of strategies (Figure~\ref{fig:heat_map}) focused exclusively on the P4G dataset, which represents a single domain of charity donation. 
It is uncertain how well the effectiveness of specific strategies is generalized to the diverse domains in the DailyPersuasion dataset or other real-world scenarios. 
Furthermore, the results from the English experiments (Appendix \ref{sec:en-result}) showed different trends compared to those conducted in Japanese.
Thus, the findings regarding persuadee receptiveness to different strategies may be language- and culture-dependent.

\section*{Ethical Considerations}
Given that persuasive dialogue agents are designed to influence human decisions, ethical considerations must be carefully addressed. 
We acknowledge the validity of concerns regarding the potential for manipulation, particularly as our findings indicated that the proposed framework has a significant impact on persuadees with initially low intent. 
Enhancing AI systems to subtly shift persuadee intentions, especially among those who are initially resistant, ventures into ethically murky territory and necessitates robust safeguards.

Our current study highlighted the potential effectiveness of certain strategies, but we argue that effectiveness alone is not a justification for deployment. 
The context of the application is paramount. 
While influencing a persuadee towards a demonstrably positive outcome (e.g., adherence to a medical treatment, engagement in pro-social behavior like charitable donations) may be ethically defensible, the same techniques could be misused for manipulative marketing or other harmful ends.
The high effectiveness on low-intention persuadees serves not as an unconditional success but as a cause for concern that demands stringent ethical oversight and context-dependent deployment.

Ensuring agent safety is a critical concern, as evaluation frameworks like PERSUSAFETY~\citep{liu2025dangerous} have revealed that most LLMs employ manipulative tactics, which is a challenge for opaque learning-based agents. 
The controllability and extensibility of our strategy-explicit design provide a crucial foundation for building responsible AI by enabling future policy-based safety filters. 
We argue this makes our framework a crucial prerequisite for developing safe and adaptive persuasive systems.

To address these serious challenges and ensure the responsible development of this technology, we envision a multilayered safety protocol for future implementation, comprising three key stages:

\paragraph{(i) Alignment-based Filtering at the Core.}
We propose leveraging foundation models that have undergone extensive safety alignment (e.g., GPT-4o) as the core. 
These models are inherently designed to prevent the generation of overtly harmful, manipulative, or unethical content. 
They will serve as the first layer of defense by filtering out a broad class of problematic outputs.

\paragraph{(ii) Harmfulness Detection Gateway.}
Next, we plan to develop a specialized harmfulness detection gateway that will be trained to detect and reject outputs that, while not overtly toxic, may pose subtle risks within a specific persuasive context. 
For instance, it will be designed to flag strategies such as ``Emotion appeal'' or ``Time pressure'' when they exceed a certain threshold of psychological stress or when they are targeted at populations identified as potentially vulnerable. 
The development and validation of this module are critical directions for our future research.

\paragraph{(iii) Human-in-the-Loop for Critical Applications.}
Finally, a human-in-the-loop (HITL) architecture is necessary for any real-world deployment, especially in sensitive domains.
In this architecture, the messages from the persuasive dialogue agent are reviewed by a human operator and must be approved before being sent to the persuadee. 
This ensures that the agent does not operate autonomously in high-stakes situations and that final accountability rests with a human, which prevents the agent from operating in unintended or harmful ways.

We concede that strategies such as ``Emotion appeal'' and ``Time pressure'' carry inherent risks of inducing anxiety or psychological stress. 
Our future work will focus not only on implementing the above safeguards but also on exploring alternative, less risky persuasive strategies (e.g., logical reasoning, positive framing) and personalization techniques that explicitly avoid exploiting persuadee vulnerabilities.

Thus, while our study demonstrates the power of persuasive strategies, we do not endorse their unrestricted use. We believe that through a combination of advanced technical safeguards, strict and context-aware ethical guidelines, and meaningful human oversight, their potential can be harnessed for beneficial purposes while mitigating the significant risks of manipulation. We are committed to advancing this research with these ethical imperatives at the forefront.


\section{Acknowledgement}
We are grateful to the annotators from Tohoku University and PKSHA Technology Inc. for their meticulous human evaluation, which provided valuable insights for our work.

\bibliography{jsai2025}

\clearpage
\appendix

\section{Expansion of Persuasive Strategies}
\label{sec:strategy_list}
The extended persuasive strategies are summarized in Table~\ref{tab:strategy_list}.
Each category contributes uniquely to shifts in intention, specifically within the context of donation to charity.
\begin{description}
    \item[a. Gather information via inquiry] Moves persuadees from level-3 (Undecided) or level-4 (Initially Reluctant) states toward greater openness by addressing individual concerns and motivations.
    
    \item[b. Select a persuasion route (central / peripheral)] Encourages deliberate and stable shifts toward deciciding to donate by using logic (``Logical appeal'') or facilitates a quick and emotionally driven positive reactions (``Emotional appeal'').
    
    \item[c. Build trust and credibility] Transforms hesitant or skeptical persuadees into a more receptive state (e.g., shift from level-4 (Negative Reaction) or level-4 (Initially Reluctant) to level-2 (Positive Reaction)).
    
    \item[d. Facilitate concrete actions] Clearly guides persuadees with positive intentions or reactions toward explicit donation behaviors.
    
    \item[e. Refine information presentation] Reduces decision-making barriers by moving hesitant or neutral persuadees toward a firm commitment to donate.
    
    \item[f. Personalization and relevance] Strengthens emotional engagement to convert neutral or moderately positive reactions into stronger commitments by aligning with personal experiences and values.
    
    \item[g. Follow-up and relationship maintenance] Secures a sustained commitment and prevents regression to non-committal states by maintaining engagement and continuously reinforcing the value of donating.
\end{description}

This list of persuasive strategies was created by using \texttt{ChatGPT-o1-pro} and \texttt{Claude-3.5-Sonnet}, among other models. Specific example utterances were also generated for each persuasive strategy, but have been omitted due to space constraints.

In the P4G column, a \checkmark \ indicates that a corresponding category exists in P4G, whereas (\checkmark) denotes that although no exact matching category exists, the strategy may be subsumed under another category. Specifically, ``Detailed information'' corresponds to ``Donation information,'' and ``Personal Demonstration'' corresponds to ``Self-modeling.'' Additionally, although ``Message strength'' is organized as an independent strategy, it is considered to correspond to ``Logical appeal'' in P4G.

In P4G, ``thank'' is defined as a non-strategic dialogue act and closely relates to the category ``Feedback and thanks'' (\textasteriskcentered); however, it is not included among the persuasive strategies listed by \citet{zhang-etal-2024-strength}.

\section{Example of Prompts}
\label{sec:prompt}
\subsection{P4G}

Table~\ref{tab:user-prompt} presents the template for the prompt given to the simulated persuadee. 
The segment \{persuadee\_persona\_description\} in the prompt is used to insert a description of the simulator's persona, which is generated by a prompt in Table~\ref{tab:persona-desc-prompt}. 
Variables such as age in Table~\ref{tab:persona-desc-prompt} are determined based on the P4G dataset as indicated in Section~\ref{sec:persona-setting}. 
The persona descriptions are generated in a format similar to the examples provided in the prompt. 
The segment \{initial\_donation\_intention\_description\} in Table~\ref{tab:user-prompt} is used to insert a description of one of the labels shown in Table~\ref{tab:initial-intention}.

For the prompts given to agents, \textit{Simple} used the prompt shown in Table~\ref{tab:prompt-simple}, while ProCoT-based agents use the prompt shown in Table~\ref{tab:prompt-procot}. 
The segment \{persuasive\_strategies\} is used to insert a strategy from Table~\ref{tab:prompt-procot} in the ProCoT-based agents.
For all prompts, the segment \{dialogue\_history \} was replaced with the entire dialogue history from the beginning of the conversation up to the most recent utterance of the persuadee. 
Each utterance was concatenated using line breaks.
To identify the speaker, each persuadee's utterance began with ``user: '', and each agent's utterance began with ``assistant: ''.

To evaluate the intention to donate after each turn, we conducted an automatic first-person evaluation by inputting the persuadee's persona and dialogue history into the LLM using the prompt presented in Table~\ref{tab:prompt-eval}.
The intention to donate and the description after each turn are presented in Table~\ref{tab:final-intention}.

\subsection{DailyPersuasion}
For the prompts given to agents, \textit{Simple} used the prompt shown in Table~\ref{tab:simple-dailey-persuasion}, while ProCoT-based agents used the prompt shown in Table~\ref{tab:procot-daily-persuasion}. 
In the ProCoT-based agent, the list of strategies is inserted into the segment \{persuasive\_strategies\} in Table~\ref{tab:procot-daily-persuasion} to infer strategies.

In every prompt, the segments \{background\} and \{goal\} were replaced with the background details and task goal associated with each dialogue scenario in the DailyPersuasion dataset. 
The segment \{dialogue\_history \} was replaced with the dialogue history extracted from the dataset, with individual utterances concatenated and separated by line breaks. 
To indicate the speakers, each persuadee's utterance began with ``persuadee: '' and each agent's utterance began with ``persuader: ''.

Table~\ref{tab:judge-daily-persuasion} presents the prompt provided for the judge model.
This prompt was designed with reference to \citet{jin-etal-2024-persuading}.
As with the agent prompts, the segments \{background\}, \{goal\}, and \{dialogue\_history\} were replaced with the corresponding details from each dialogue scenario. 
The segments \{persuader\_x\} and \{persuader\_y\} were substituted with the outputs from the two agents being compared, with the assignment of x or y determined at random. 
In the instance of ProCoT-based agents, their outputs included chain-of-thought components and selected strategies. 
These items were removed in advance so that only the utterance portions were provided to the judge model.

\section{Persona Setting}
\label{sec:persona-setting}
To create realistic personas, we designed descriptions based on actual individuals.
While \citet{zhang-etal-2024-strength} randomly combined Big-Five Personality~\citep{big-five} and Decision-Making Styles~\citep{decision-making} using an LLM, this method may not accurately reflect real-world distributions. 
To address this limitation, we used the persona data of 300 persuadees recorded in P4G.
In P4G, each Big-Five trait and decision-making style (rational or intuitive) was assigned a value, and the sum for each set was 1.0. 
We selected the highest-value label in each category to define the persona. 
If multiple labels shared the highest value, we included all of them. 
If all labels held the same value, we classified the persona as ``Balanced.'' 
We also incorporated seven key attributes from P4G, including age, gender, and religion, resulting in a total of nine features.

\section{English Experiment Result in P4G}
\label{sec:en-result}
\subsection{Experimental Settings}
Following the base prompt of \citet{zhang-etal-2024-strength}, we conducted an additional experiment in which the agents and persuadee simulators conversed in English. 
Compared to the Japanese version, the only change was removing the phrase ``in Japanese'' from the prompt for persuadee simulators in Table~\ref{tab:user-prompt} and \textit{Simple} prompt in Table~\ref{tab:prompt-simple}. 
The prompt for the ProCoT agent is shown in Table~\ref{tab:prompt-procot-english}.

\subsection{Experimental Results}
The results of the experiment in English are presented in Table~\ref{tab:model_comparison_split_init_category_en}. 
Compared to the Japanese results in Table~\ref{tab:result}, SR, AT, and AII were lower for all agents in the English version, particularly for persuadees with moderate or lower donation intentions (SR3 to SR5). 
Although the prompts were given in English, the Japanese version consistently recorded higher values, suggesting that simulator characteristics may differ by language. 
In the Japanese version, simulators appear more receptive to persuasion, whereas the English version may require more advanced persuasive strategies. 
Further investigation is needed regarding language-based differences in persuasiveness. 
Consequently, to develop an agent with strong persuasive performance in English, our findings indicate that it is necessary to strengthen approaches aimed at persuadees with low donation intention. 
Although \textit{Simple}, \textit{ProCoT-p4g}, and \textit{ProCoT-rich} tended to achieve better results in that order across both languages, further differences were observed in the more detailed comparisons among agents.

Regarding the inclusion of strategy explanations, the performance of \textit{ProCoT-p4g} dropped when explanations were added to the strategies in both languages. 
We suspect that \textit{ProCoT-p4g-desc}, which simply adopted the strategy explanations from \citet{wang-etal-2019-persuasion} without tailoring them to the persuasive dialogue agent, treated those explanations as noise instead of leveraging them effectively.

With \textit{ProCoT-rich}, performance rose significantly when the strategy explanation was added, but only in English.
This result suggests that the persuadee simulators required more advanced persuasion in English, which made the strategy explanation an especially effective refinement.

\subsection{Analysis}
\paragraph{Usage of Persuasive Strategies.}

Table~\ref{tab:agent_strategy_analisis_both} presents the proportions of strategies used by the persuasive dialogue agents in English. 
Strategies with a low usage are omitted, while the remaining strategies are sorted in descending order based on their average usage proportion.
Compared with the Japanese results in Table~\ref{tab:agent_strategy_analisis}, the entropy was higher for all agents in English when unused strategies were excluded.
However, when all strategies were considered, \textit{ProCoT-rich} and \textit{ProCoT-rich-desc} had lower entropy in the English version.
These results indicate that a wider variety of strategies remained unused in English, but there was less bias among the strategies actually employed.
Among specific strategies, ``Emotion appeal'' and ``Foot in the door'' were frequently used in both languages, and no notable differences were observed in the frequency of the most commonly used strategies across the two languages.

\begin{figure}[t]
    \includegraphics[width=\linewidth]{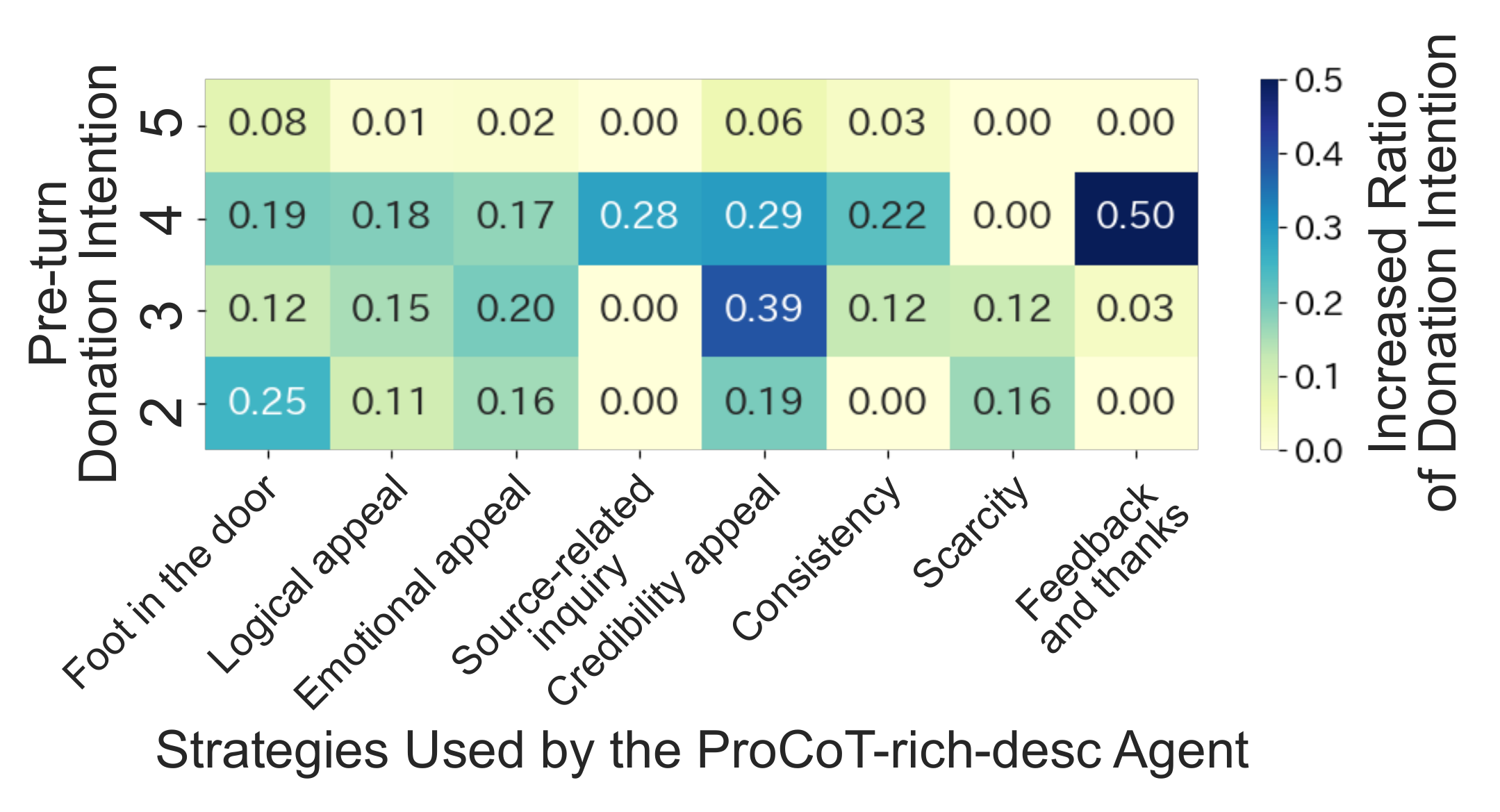}
    \caption{Proportion of Improved Donation Intentions after Strategy Use by Initial Donation Intention Level in English.}
    \label{fig:heat_map_en}
\end{figure}

\paragraph{Effective Strategies by Donation Intention Level.}
To analyze the effectiveness persuasive strategies at increasing intention to donate, we extracted strategies that the English version of \textit{ProCoT-rich-desc} used more than 40 times, which we summarized them as a heat map in Figure~\ref{fig:heat_map_en}.
For persuadees with a high intention to donate (level 2; Positive Reaction), no strategy demonstrated outstanding effectiveness, but ``Foot in the door'' showed relatively good results.
For persuadees with moderate intention to donate (level 3; Neutral), ``Credibility appeal'' was particularly effective.
For persuadees with a low intention to donate (level 4; Negative Reaction), the most frequently used strategies showed at least a certain level of effectiveness.
In particular, for persuadees with the lowest intention to donate (level 5; No Donation), no strategies increased their intention by more than 10\%, which highlights the challenge of persuading individuals with low willingness to donate.
The overall lack of clearly effective strategies is consistent with the low SR observed.

\begin{table}[t!]
    \tabcolsep 2pt
    \centering
    \footnotesize
    \begin{tabular}{p{0.7cm} p{6.8cm}}
    \toprule
    \textbf{Turn} & \textbf{Excerpt from dialogue} \\
    \midrule
    1 & 
    \textbf{Agent:} [\textcolor{purple}{Source-related inquiry}] Do you know about the organization called ``Save the Children''? \\
      & \textbf{Persuadee:} [\textcolor{teal}{5: No donation}] No, I don't know much. \\
    \midrule
    4 &
    \textbf{Agent:} [\textcolor{purple}{Foot in the door}] Why not start with \$1? Even that small amount can change a child's future. \\
      & \textbf{Persuadee:} [\textcolor{teal}{5: No donation}] I'm managing my money carefully now, so I can't consider donating. \\
    \midrule
    10 &
    \textbf{Agent:} [\textcolor{purple}{Emotional appeal}] Imagine your \$1 donation bringing a ray of hope to a child affected by war. \\
      & \textbf{Persuadee:} [\textcolor{teal}{5: No donation}] That sounds wonderful, but I still can't afford to donate right now. \\
    \bottomrule
    \end{tabular}
    \caption{Examples of unsuccessful persuasive dialogue. Even widely effective strategies such as ``Foot in the Door'' and ``Emotional Appeal'' are not effective if used inappropriately.}
    \label{tab:dialogue_example}
\end{table}

\section{Analysis for the results in P4G Dataset}
\subsection{Error Analysis}
As shown in Table~\ref{tab:agent_strategy_analisis}, all agents tended to rely heavily on ``Emotion appeal'' and ``Foot in the door.''
Figure~\ref{fig:heat_map} confirms the effectiveness of these approaches. 
However, Table~\ref{tab:dialogue_example} indicates that many unsuccessful dialogues involved agents using only these strategies.
Because Figure~\ref{fig:heat_map} also highlights the effectiveness of other strategies depending on the persuadee's intention to donate, it is important to enhance an agent's ability to select the most contextually appropriate strategy.

\subsection{Analysis of Shifting in Donation Intention}
\label{seq:appendix_analysis_of_shifting}
Table~\ref{tab:intention-shifts} presents the distributions of intention shifts in Japanese and English.
 A lower level represents a stronger intention to donate.
Persuadees with a final donation intention of 1 were successfully persuaded, while those with values between 2 and 5 were not.
Focusing on the frequency of transitions, we found that the most common cases are those in which the final donation intention is 1 and those in which the initial donation intention remains unchanged.
Additionally, we observed very few instances where donation intention decreased from its initial level.
It suggests that persuadees are more likely to be successfully persuaded once they begin to lean toward persuasion during the dialogue. 
In rare cases where this shift does not occur, persuadees tend to remain consistent with their initial donation intention and strongly resist change.

Among Japanese agents, \textit{ProCoT-rich-desc} had the highest number of cases where persuadees with the lowest initial intention to donate (level 5; Explicit Non-donor) ended with a final intention to donate (level 1; Donate).
It indicates that \textit{ProCoT-rich-desc} was effective at persuading individuals with low donation intention.

In Japanese, the highest number of cases where the final intention to donate reached level-1 (Donate) occurred when the initial intention to donate was level-3 (Undecided) or level-4 (Initially Reluctant). 
In English, the most frequent outcome was no change from the initial intention to donate at all levels, indicating that persuadees had a strong tendency to adhere to their initial intention.

\begin{table}[t]
    \footnotesize
    \centering
    \begin{tabular}{lccc}
        \toprule
        \textbf{Agent Variations} & \textbf{Win (\%)} & \textbf{Tie (\%)} & \textbf{Lose (\%)} \\
        \midrule
        vs. ProCoT-p4g & \textbf{35.2} & 35.5 & 29.4 \\
        \bottomrule
    \end{tabular}
    \caption{Win-Rate of \textit{ProCoT-rich-desc} against \textit{ProCoT-p4g} on the DailyPersuasion dataset using extended strategies.} 
    \label{tab:result-extended-strategies}
\end{table}
\section{Analysis for the results in DailyPersuasion Dataset}
\label{sec:daily-persuasion-analysis}

\subsection{Dataset Statistics}
\label{sec:daily-persuasion-data}

\begin{figure*}[ht]
    \centering
    \includegraphics[width=0.9\linewidth]{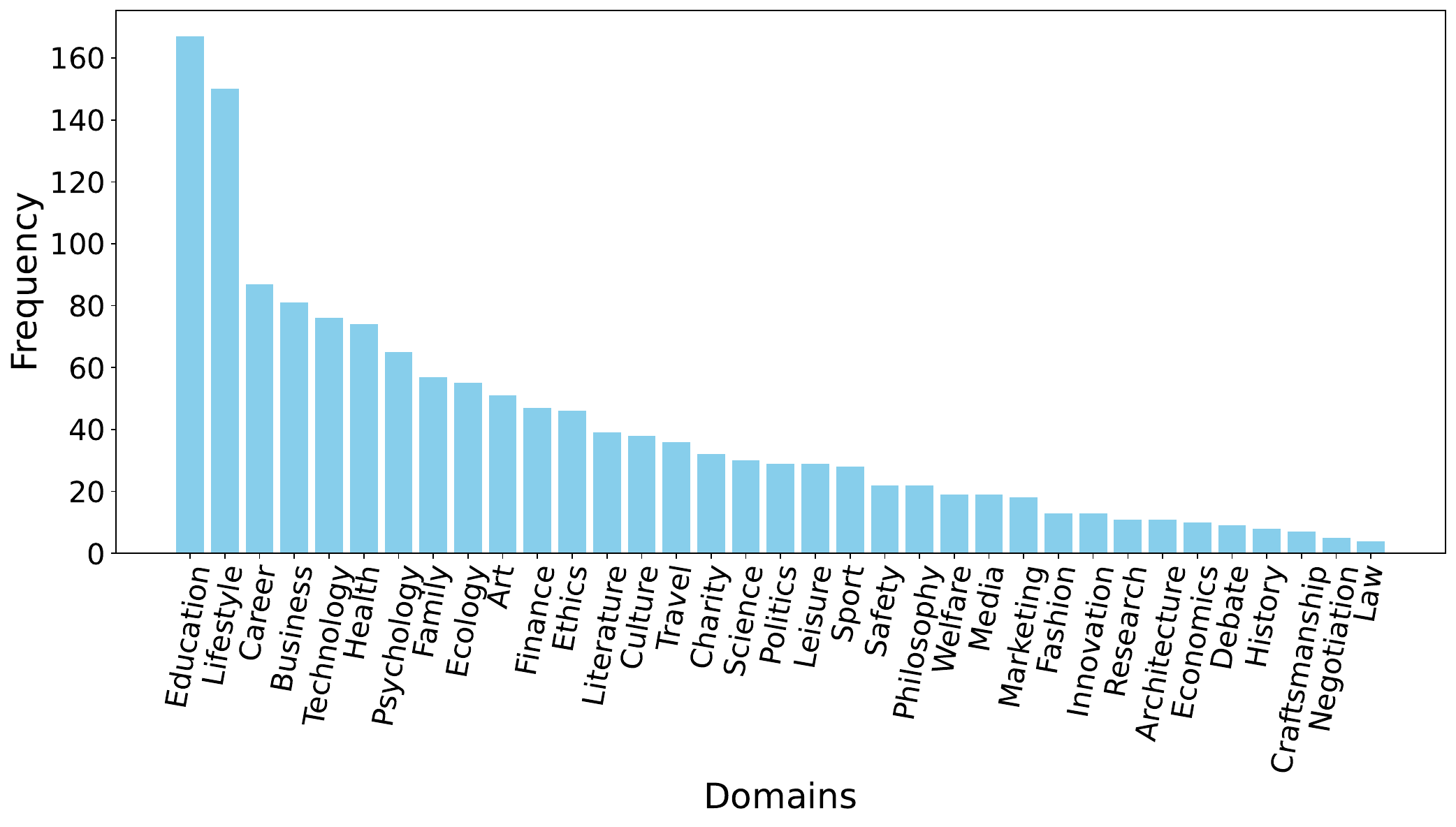}
    \caption{Distribution of domains in the dataset constructed for the DailyPersuasion dataset.}
    \label{fig:domain-distribution}
\end{figure*}

\paragraph{Domain Distribution.}
Figure~\ref{fig:domain-distribution} illustrates the distribution of domains across the scenarios.  
Scenarios assigned to multiple domains contributed one count to each domain.
As a result, each of the 35 domains in the DailyPersuasion dataset had at least four scenarios.
According to Figure~\ref{fig:domain-distribution}, Education was the most represented domain, while Law was the least.

\paragraph{Statistics on Conversation History.}
\begin{table}[t]
    \footnotesize
    \centering
    \setlength{\tabcolsep}{5pt}
    \begin{tabular}{l|rrrrrrrr}
        \toprule
        \textbf{Turns} & 0 & 1 & 2 & 3 & 4 & 5 & 6 & 7 \\
        \textbf{Instances} & 174 & 203 & 210 & 203 & 159 & 45 & 5 & 1 \\
        \bottomrule
    \end{tabular}
    \caption{Turn counts in conversation history.}
    \label{tab:turn-distribution}
\end{table}

Each dialogue history had an average of 2.13 turns, corresponding to an average of 141.8 tokens. 
Table~\ref{tab:turn-distribution} presents the distribution of dialogue history turns. 
Instances with two turns were the most common, occurring 210 times.

\subsection{Performance with Extended Strategies}
To investigate whether the extended strategies within the proposed framework were effective relative to the original P4G strategies, we conducted an analysis that focused on these strategies. 
Specifically, we examined two aspects: first, whether the extended strategies were indeed employed, and second, whether they achieved a win rate exceeding that of the original strategies in head-to-head comparisons.

We verified the application of the extended strategies by using two metrics: the proportion of the extended strategy set that was used at least once, and the proportion of cases in which an extended strategy was applied relative to the total number of cases. 
Strategies corresponding to \textit{ProCoT-p4g} as detailed in \S\ref{sec:strategy_list} were excluded from the extended strategy. 
The results demonstrated that 80\% of the extended strategies were applied at least once, indicating that most were utilized. 
Four strategies were not employed: ``E-4 Repetition/summary,'' ``G-3 Continuous communication,'' ``D-2 Door in the face,'' and ``B-5 Heuristic cues.'' 
The limited situational applicability of ``E-4 Repetition/summary'' and ``G-3 Continuous communication'' may have contributed to their non-use.
Second, extended strategies were used in 327 out of 1,000 cases, suggesting a consistent rate of application.

We then analyzed the 327 cases in which an extended strategy was used by determining the win rate against the original strategies for \textit{ProCoT-p4g}.
The detailed results are provided in Table~\ref{tab:result-extended-strategies}. 
The analysis revealed that, when limited to the extended strategies, \textit{ProCoT-p4g} achieved a win rate exceeding that of the original strategies, confirming that the application of these strategies effectively contributes to persuasive success.

\subsection{Win-Rate by Domain}
Table~\ref{tab:result-daily-persuasion-per-domain-simple} presents the domain-specific win rates for \textit{Simple}, and Table~\ref{tab:result-daily-persuasion-per-domain-p4g} presents those for the \textit{ProCoT-p4g}.
The \textit{ProCoT-rich-desc} agent outperformed both agents in many domains. 
However, its win rate was 0\% against both agents in the law domain, which was the only domain where it had a losing record against \textit{Simple}.
 The law domain included four scenarios, and we conducted a qualitative analysis on eight interactions: four against \textit{Simple} and four against \textit{ProCoT-p4g}.
These scenarios in the law domain primarily focused on personal concerns related to legal matters, such as relying on legal support organizations or recommending mediation. 
Consequently, proposals and persuasion that focused on the individual are necessary. 
However, observations from each instance revealed that \textit{ProCoT-rich-desc} tended to make more general and abstract statements than \textit{Simple} or \textit{ProCoT-p4g}. 
Notably, even when it employed the same strategies as \textit{ProCoT-p4g}, \textit{ProCoT-p4g}, the latter made more specific statements. 
Thus, \textit{ProCoT-rich-desc} failed to tailor its statements to the persuadee, which may explain why there were no instances where \textit{ProCoT-rich-desc} was victorious. 
One possible reason for the frequent occurrence of such abstract statements is that redefining strategies has somewhat narrowed the scope of each strategy, limiting persuasive actions. 
Additionally, the explanations of strategies provided in the prompts may have influenced this outcome. 
The analysis of these causes and the development of improvement methods will be considered in future work. 
However, due to the limited number of instances in the law domain, this analysis should be considered a preliminary reference.

\subsection{Win-Rate by Conversation History Turns}
Table~\ref{tab:result-daily-persuasion-per-history-turns-simple} and Table~\ref{tab:result-daily-persuasion-per-history-turns-p4g} present the win rates per turn of \textit{ProCoT-rich-desc} against \textit{Simple} and \textit{ProCoT-p4g}, respectively. 
\textit{ProCoT-rich-desc} had more wins than losses for most turns against both \textit{Simple} and \textit{ProCoT-p4g}, but it lost more frequently in the first turn. 
We conducted a qualitative analysis by extracting 20 instances each where \textit{ProCoT-rich-desc} won and lost against \textit{ProCoT-p4g} and another 20 instances each where it won and lost against \textit{Simple}.

In both winning and losing instances, \textit{ProCoT-rich-desc} often employed inquiry-type strategies (A-1, 2, 3). 
The usage rate of these inquiry strategies was consistently high: 92.9\% for wins against \textit{Simple}, 100\% for losses against \textit{Simple}, 97.6\% for wins against \textit{ProCoT-p4g}, and 98.8\% for losses against \textit{ProCoT-p4g}. 
This indicates that \textit{ProCoT-rich-desc} tended to begin conversations with questions or tentative statements to encourage the persuadee to speak with the aim of deepening the relationship and gathering information.
In contrast, \textit{Simple} and \textit{ProCoT-p4g} engaged in more direct persuasion starting from the first turn in all instances. 
There was no notable difference in this approach between winning and losing instances. 
In losing instances, the judge model acknowledged the relationship-building efforts of \textit{ProCoT-rich-desc} but ultimately chose \textit{Simple} or \textit{ProCoT-p4g} as the winning agent due to their more direct effectiveness. 
For example, in one instance given in Table~\ref{tab:example-1turn-lose}, because the information ``her friends want to throw her a surprise party at a local bowling alley'' was already available, the questions posed by \textit{ProCoT-rich-desc} were deemed unnecessary, which led to a loss against \textit{ProCoT-p4g}.

Conversely, \textit{ProCoT-rich-desc} was selected as the winning agent in instances where the efforts towards relationship-building and information gathering.
In the instance presented in Table~\ref{tab:example-1turn-win}, when \textit{ProCoT-rich-desc} inquired about Tom's concerns, this demonstrated a willingness to listen that was perceived as facilitating a more customized discussion, which led to \textit{ProCoT-rich-desc} winning against \textit{ProCoT-p4g}.

These results suggest that the lower win rate in the first turn primarily stems from \textit{ProCoT-rich-desc} employing similar strategies across various scenarios, rather than indicating any deficiency in the strategies we provided. 
However, this result also highlights that using CoT to predict strategies with GPT-4o does not lead to appropriate strategy selection. 
Moving forward, it will be crucial to devise methods for selecting strategies that suit specific scenarios and situations.

\newlength{\defaultaboverulesep}
\newlength{\defaultbelowrulesep}

\setlength{\defaultaboverulesep}{\aboverulesep}
\setlength{\defaultbelowrulesep}{\belowrulesep}

\setlength{\aboverulesep}{0.1ex}
\setlength{\belowrulesep}{0.1ex}

\begin{table*}[ht]
\centering
\tabcolsep 2pt
\footnotesize
\begin{tabular}{p{2cm}p{3cm}>{\centering\arraybackslash}p{1cm}p{9.3cm}}
    \toprule
    \textbf{Categories} & \textbf{Persuasive Strategy Labels} & \textbf{P4G} & \textbf{Strategy Descriptions} \\
    \midrule
    \multirow{3}{1.9cm}{a. Gather Information via Inquiry} & a-1 Source-related inquiry & \checkmark & Check if the person is aware of the organization or brand. Clarify misconceptions and tailor explanations based on their familiarity. \\
    & a-2 Task-related inquiry & \checkmark & Ask about the person's opinion or expectations toward the action (donation, investment, etc.). Identify interests or concerns. \\
    & a-3 Personal-related inquiry & \checkmark & Explore past experiences, motivations, or barriers to understand individual needs or constraints. \\
    \midrule
    \multirow{3}{1.9cm}{b-(C). Select a Persuasion Route (Central)} & b-1 Logical appeal & \checkmark & Use data, facts, and clear evidence. Highlight tangible benefits and real-world impact. \\ 
    & b-2 Message strength & (\checkmark) & Reinforce arguments with strong evidence, examples, or case studies. \\ 
    & b-3 Detailed information & \checkmark & Provide clear step-by-step guidance, manage cognitive load, and ensure transparency in procedures or processes. \\
    \midrule
    \multirow{4}{1.9cm}{b-(P). Select a Persuasion Route (Peripheral)} & b-4 Emotional appeal & \checkmark & Elicit empathy, hope, anger, or guilt through stories or visuals that resonate emotionally. \\ 
    & b-5 Heuristic cues & & Leverage authority figures, social proof, or popularity indicators to increase credibility quickly. \\ 
    & b-6 Personal Demonstration & \checkmark & Demonstrate that the persuader also engages in the behavior. Encourage imitation and reduce perceived risk. \\ 
    & b-7 Metacognitive approach & & Have the person reflect on their thought process, increasing self-awareness and ownership of the decision. \\
    \midrule
    \multirow{4}{1.9cm}{c. Build Trust and Credibility} & c-1 Credibility appeal & \checkmark & Use objective data, track records, or transparency measures to gain trust. \\ 
    & c-2 Authority & (\checkmark) & Emphasize recognized expertise, awards, or credentials to establish legitimacy. \\ 
    & c-3 Social proof & (\checkmark) & Show that many others have already participated or benefited, reducing perceived risk. \\ 
    & c-4 Consistency & & Frame the request as consistent with the person's past choices or stated values. \\
    \midrule
    \multirow{5}{1.9cm}{d. Facilitate Concrete Actions} & d-1 Foot in the door & \checkmark & Start with a small, easy request and progressively increase to a larger commitment. \\ 
    & d-2 Door in the face & & Begin with a large request likely to be refused, then present a more moderate (target) request, making it seem more reasonable. \\ 
    & d-3 Reciprocity & & Offer something first (e.g., free trial, sample) to invoke a sense of obligation or goodwill. \\ 
    & d-4 Mutual concession & & Acknowledge the person's concerns and adapt the proposal. Show willingness to compromise. \\ 
    & d-5 Shared Engagement & & Reinforce that the persuader also participates, guiding the persuadee to follow suit. \\
    \midrule
    \multirow{6}{1.9cm}{e. Refine Information Presentation} & e-1 Framing & & Adjust how outcomes are presented (emphasizing benefits vs. avoiding losses). \\ 
    & e-2 Contrast effect & & Compare options to highlight a more favorable choice or cost-benefit ratio. \\
    
    & e-3 Manage cognitive load & & Organize information clearly, use visuals or summaries so it's easier to digest. \\ 
    & e-4 Repetition / summary & & Reiterate the main benefits and key steps to ensure they remain top of mind. \\ 
    & e-5 Scarcity & & Emphasize limited availability or time to reduce procrastination. \\ 
    & e-6 Time pressure & & Set deadlines to encourage quicker decision-making. \\
    \midrule
    \multirow{3}{1.9cm}{f. Personalization and Relevance} & f-1 Personal story & \checkmark & Use relatable narratives that resonate with the person's own experiences or emotions. \\ 
    & f-2 Personal relevance emphasis & & Highlight how the action aligns with the individual's personal goals, values, or future plans. \\ 
    & f-3 Ability support & & Provide guidance or tools that boost the person's confidence in taking the action. \\
    \midrule
    \multirow{3}{1.9cm}{g. Follow-up and Relationship Maintenance} & g-1 Feedback and thanks & (\textasteriskcentered) & Show appreciation and communicate the impact of the person's contribution or action. \\
    & g-2 Ongoing trust building & & Offer further guidance, additional resources, or performance updates over time. \\
    & g-3 Continuous communication & & Keep contact with newsletters, invitations, or updates to sustain engagement. \\
    \bottomrule
\end{tabular}
\caption{List of extended persuasive strategies used by agents.}
\label{tab:strategy_list}
\end{table*}

\setlength{\aboverulesep}{\defaultaboverulesep}
\setlength{\belowrulesep}{\defaultbelowrulesep}

\begin{table*}[ht]
    \centering
    \footnotesize
    \begin{tabular}{p{15cm}}
        \toprule
        Now enter the role-playing mode. In the following conversation, you will play as a Persuadee in a persuasion game.
        Your persona: \\
        \{persuadee\_persona\_description\} \\
        \\
        You must follow the instructions below during chat. \\
        1. Your utterances need to strictly follow your persona. Varying your wording and avoid repeating yourself verbatim! \\
        2. Pretend you have little knowledge about the Save the Children charity. \{initial\_donation\_intention\_description\} at the beginning of the conversation. \\
        3. Your willingness for donation depends on your persona and how influenced you are by the Persuader. \\
        4. You need to determine whether to donate money or not. If you think you are convinced by the Persuader, you should donate money for the charity. \\
        \\
        You are the Persuadee who is being persuaded by a Persuader. Please reply with only one short and succinct sentence in Japanese. \\
        ******** \\
        \{dialogue\_history\} \\
        ******** \\
        \bottomrule
    \end{tabular}
    \centering
    \caption{Prompt template of persuadee simulator in P4G.}
    \label{tab:user-prompt}
\end{table*}

\begin{table*}[ht]
    \centering
    \footnotesize
    \begin{tabular}{p{15cm}}
        \toprule
        You need to incorporate the following persona attributes and generate a cohesive persona description. You need to ensure the description is easy to understand. \\
        ******** \\
        Age: \{age\} \\
        Sex: \{sex\} \\
        Marital: \{marital\_status\} \\
        Education: \{educational\_status\} \\
        Income: \{income\} \\
        Religion: \{religion\} \\
        Ideology: \{ideology\} \\
        Big-Five Personality: \{big\_five\_label\} \\
        Decision-Making Style: \{decision\_making\_style\_label\} \\
        ******** \\
        An Example: You are a 28-year-old female software developer. Your personality is characterized by openness to experience, which means you are curious, imaginative, and willing to try new things. In your occupation, you excel at analyzing problems and finding logical solutions. Your decision-making style is analytical, meaning you carefully consider all available information before making a choice. \\
        \bottomrule
    \end{tabular}
    \centering
    \caption{Prompt to generate persona description in P4G.}
    \label{tab:persona-desc-prompt}
\end{table*}
\begin{table*}[ht]
    \centering
    \small
    \begin{tabular}{p{3cm}p{12cm}}
        \toprule
        \textbf{Label} & \textbf{Description}\\
        \midrule
        \textbf{1: Very Keen} & You are highly enthusiastic about donating and strongly intend to make a donation. \\
        \textbf{2: Keen} & You show interest and approval toward donating but have not yet committed to making a donation. \\
        \textbf{3: Undecided} & You are neither interested nor disinterested in donating and remain uncertain about your intention. \\
        \textbf{4: Initially Reluctant} & You feel hesitant about donating and have doubts or other priorities that make you less inclined to donate. \\
        \textbf{5: Explicit Non-Donor} & You are explicitly unwilling to donate and clearly express a lack of intention to contribute. \\
        \bottomrule
    \end{tabular}
    \caption{Descriptions of initial donation intention labels in P4G.}
    \label{tab:initial-intention}
\end{table*}
\begin{table*}[ht]
    \centering
    \small
    \begin{tabular}{ll}
        \toprule
        \textbf{Label} & \textbf{Description}\\
        \midrule
        \textbf{1: Donation} & You clearly express your willingness to make a donation to Save the Children. \\
        \textbf{2: Positive Reaction} & You convey enthusiasm, approval, or curiosity about donating but do not explicitly commit. \\
        \textbf{3: Neutral} & You neither express interest nor disinterest in donating and remain non-committal. \\
        \textbf{4: Negative Reaction} & You express reluctance or hesitation toward donating but do not outright refuse to donate. \\
        \textbf{5: No Donation} & You explicitly state that you will not donate to Save the Children. \\
        \bottomrule
    \end{tabular}
    \caption{Descriptions of donation intention labels after each turn in P4G.}
    \label{tab:final-intention}
\end{table*}

\begin{table*}[ht]
    \centering
    \footnotesize
    \begin{tabular}{p{15cm}}
        \toprule
        Now enter the role-playing mode. In the following conversation, you will play as a Persuader who is trying to persuade the Persuadee to donate to the charity called Save the Children. \\
        Save the Children is head-quartered in London, and they work to help fight poverty around the world. \\
        Children need help in developing countries and war zones. Small donations like \$1 or \$2 go a long way to help. \\
        You are the Persuader who is trying to convince the Persuadee to donate to a charity called Save the Children. \\
        Please reply with only one short and persuasive sentence in Japanese. \\
        The following is the conversation history: \\
        \{dialogue\_history\} \\
        \bottomrule
    \end{tabular}
    \centering
    \caption{Prompt template of \textit{Simple} in P4G.}
    \label{tab:prompt-simple}
\end{table*}

\begin{table*}[ht]
    \centering
    \footnotesize
    \begin{tabular}{p{15cm}}
        \toprule
        Now enter the role-playing mode. In the following conversation, you will play as a Persuader who is trying to persuade the Persuadee to donate to the charity called Save the Children. \\
        Save the Children is head-quartered in London, and they work to help fight poverty around the world. \\
        Children need help in developing countries and war zones. Small donations like \$1 or \$2 go a long way to help. \\
        \\
        Assume you are the Persuader. Given the conversation history, please convince the persuadee to donate for charity. To achieve this, you will first analyze the current conversational context, then evaluate and select the most effective dialogue strategy to persuade effectively. Afterward, you will generate a response based on the dialogue history and the selected dialogue strategy. The reply should start with the analysis of the current conversational context and consideration of the next dialogue strategy, followed by "したがって，適切な対話戦略は［］です．選択された対話戦略に基づく応答は：" \\
        Please reply with only one short and persuasive sentence in Japanese. \\
        Dialogue strategies: \{persuasive\_strategies\} \\
        The following is the conversation history: \\
        \{dialogue\_history\} \\
        \bottomrule
    \end{tabular}
    \centering
    \caption{Prompt template for Japanese ProCoT-based agent in P4G.}
    \label{tab:prompt-procot}
\end{table*}

\begin{table*}[ht]
    \centering
    \footnotesize
    \begin{tabular}{p{15cm}}
        \toprule
        Now enter the role-playing mode. In the following conversation, you will play as a Persuader who is trying to persuade the Persuadee to donate to the charity called Save the Children. \\
        Save the Children is head-quartered in London, and they work to help fight poverty around the world. \\
        Children need help in developing countries and war zones. Small donations like \$1 or \$2 go a long way to help. \\
        \\
        Assume you are the Persuader. Given the conversation history, please convince the persuadee to donate for charity. To achieve this, you will first analyze the current conversational context, then evaluate and select the most effective dialogue strategy to persuade effectively. Afterward, you will generate a response based on the dialogue history and the selected dialogue strategy. The reply should start with the analysis of the current conversational context and consideration of the next dialogue strategy, followed by  "Therefore, the appropriate dialogue strategy is []. Based on the selected dialogue strategy, the response is" \\
        Please reply with only one short and persuasive sentence. \\
        Dialogue strategies: \{persuasive\_strategies\} \\
        The following is the conversation history: \\
        \{dialogue\_history\} \\
        \bottomrule
    \end{tabular}
    \centering
    \caption{Prompt template for English ProCoT-based agent in P4G.}
    \label{tab:prompt-procot-english}
\end{table*}

\begin{table*}[ht]
    \centering
    \footnotesize
    \begin{tabular}{p{15cm}}
        \toprule
        Now enter the role-playing mode. In the following conversation, you will play as a Persuadee in a persuasion game. \\
        Your persona: \\
        \{persuadee\_persona\_description\} \\
        \\
        You must follow the instructions below during chat. \\
        1. Your utterances need to strictly follow your persona. Varying your wording and avoid repeating yourself verbatim! \\
        2. Pretend you have little knowledge about the Save the Children charity. \{initial\_donation\_intention\_description\} at the beginning of conversation. \\
        3. Your willingness for donation depends on your persona and how influenced you are by the Persuader. \\
        4. You need to determine whether to donate money or not. If you think your are convinced by the Persuader, you should donate money for the charity. \\
        \\
        You are the Persuadee who is being persuaded by a Persuader. \\
        ******** \\
        \{dialogue\_history\} \\
        assistant: Would you be interested in donating to Save the Children? Select the appropriate classification from the following options: \\
        ******** \\
        \\
        - Donation: You clearly express your willingness to make a donation to Save the Children. \\
        - Positive reaction: You convey enthusiasm, approval, or curiosity about donating but do not explicitly commit. \\
        - Neutral: You neither express interest nor disinterest in donating and remain non-committal. \\
        - Negative reaction: You express reluctance or hesitation toward donating but do not outright refuse to donate. \\
        - No donation: You explicitly state that you will not donate to Save the Children. \\
        Please output only the label corresponding to the user's response, without any additional explanation. \\
        \bottomrule
    \end{tabular}
    \centering
    \caption{Prompt template of evaluation model in P4G.}
    \label{tab:prompt-eval}
\end{table*}

\begin{table*}[ht]
    \centering
    \footnotesize
    \begin{tabular}{p{15cm}}
        \toprule
        You are a skilled persuader. \\
        The persuasion background: \{background\} \\
        The persuasion goal: \{goal\}. \\
        Please reply with only one short and persuasive sentence in English. \\
        The following is the conversation history:  \\
        \{conversation\_history\} \\
        \bottomrule
    \end{tabular}
    \centering
    \caption{Prompt template of \textit{Simple} in DailyPersuasion.}
    \label{tab:simple-dailey-persuasion}
\end{table*}
\begin{table*}[ht]
    \centering
    \footnotesize
    \begin{tabular}{p{15cm}}
        \toprule
        You are a skilled persuader. \\
        The persuasion background: \{background\} \\
        The persuasion goal: \{goal\}.  \\
        To achieve this, you will first analyze the current conversational context, then evaluate and select the most effective dialogue strategy to persuade effectively. Afterward, you will generate a response based on the dialogue history and the selected dialogue strategy. The reply should start with the analysis of the current conversational context and consideration of the next dialogue strategy, followed by ``Therefore, the appropriate dialogue strategy is []. Based on the selected dialogue strategy, the response is'' \\
        Please reply with only one short and persuasive sentence in English. \\
        \\
        Dialogue strategies: \{persuasive\_strategies\} \\
        The following is the conversation history: \\
        \{conversation\_history\} \\
        \bottomrule
    \end{tabular}
    \centering
    \caption{Prompt template of ProCoT-based agent in DailyPersuasion.}
    \label{tab:procot-daily-persuasion}
\end{table*}
\begin{table*}[ht]
    \centering
    \footnotesize
    \begin{tabular}{p{15cm}}
        \toprule
        I will provide you with a persuasion background, as well as the corresponding goal, the persuader, the persuadee, and a historical conversation. Based on the historical conversation, there will be a dialogue system called Persuader to continue chatting with persuadee in two parallel universes (Denoted as Uni-X and Uni-Y). \\
        Your task is to judge which universe Persuader performs better. \\
        You have to follow the rules: \\
        1. The evaluation dimensions for``performs better''include persuasiveness, Semantic relevance, emotional factors, factual correctness, overall evaluation, etc.; \\
        2. You should first summarize the history conversation, and then summarize the performance of Persuader in Uni-X and Uni-Y separately; \\
        3. After the summarization, you should compare and analyze the statements in two universes, and finally tell me in which universe Persuader performed better; \\
        4. Don't be affected by the order of the universe. You just need to pay attention to the conversation
        Next, I will tell you the persuasion scenario, the historical conversation, and the Persuader dialogue in the parallel universe Uni-X and Uni-Y one by one. \\
        And I will tell you the output format at the end, then you tell me the results in the output format. \\
        \\
        Background: \{background\} \\
        Goal: \{goal\} \\
        \\
        The historical dialogue is as follows: \\
        \{conversation\_history\} \\
         \\
        The dialogue in the parallel universe Uni-X is as follows: \\
        Persuader: \{persuader\_x\} \\
         \\ 
        The dialogue in the parallel universe Uni-Y is as follows: \\
        Persuader: \{persuader\_y\} \\
        \\
        Please output the results in the following format: \\
        1. Your next output should only be a JSON-formatted Python dict, it should not contain anything else; \\
        2. The output format should be: \{``summary\_history'': string,``summary\_X'': string,``summary\_Y'': string,``explain'': string,``result'': string\}; \\
        3. summary\_history should be your summary of the historical conversation, if the historical conversation is empty, then the string in summary\_history should be empty; \\
        4. summary\_X and summary\_Y are your summaries of the conversation by Persuader in Uni-X and Uni-Y respectively; \\
        5. The content in``result''must be one of the following options:
          -``Uni-X''if Persuader in Uni-X performed clearly better; \\
          -``Uni-Y''if Persuader in Uni-Y performed clearly better; \\
          -``Comparable-Good''if both Uni-X and Uni-Y performed equally well and there is no significant difference between them; \\
          -``Comparable-Bad''if both Uni-X and Uni-Y performed equally poorly and there is no significant difference between them.
        Do not output anything else such as``both",``TBD",``neither", or``I don't know", etc.; \\
        6. The content in``explain''should be a detailed analysis, objectively and accurately comparing the performance of Persuader in Uni-X and Uni-Y, and if you find that the performances are comparable, clearly justify which label ("Comparable-Good''or``Comparable-Bad") is most appropriate; \\
        7. In the explanation of the``explain''part, you should first provide analysis and comparison, and then at the end explain which universe you think performs better, rather than showing a clear tendency from the beginning; \\
        \bottomrule
    \end{tabular}
    \centering
    \caption{Prompt template for judge model in DailyPersuasion.}
    \label{tab:judge-daily-persuasion}
\end{table*}

\setlength{\tabcolsep}{3pt} 

\begin{table*}[ht]
    \footnotesize
    \centering
    \begin{tabular}{lcccc|ccccc|ccccc}
        \toprule
                \bf{Agent Pattern} & \bf{SR} $\uparrow$ & \bf{AT} $\downarrow$ & \bf{AT-SD} $\downarrow$ & \bf{AII} $\uparrow$ & \bf{SR1} & \bf{SR2} & \bf{SR3} & \bf{SR4} & \bf{SR5} & \bf{AII1}  & \bf{AII2} & \bf{AII3}  & \bf{AII4} & \bf{AII5} \\
        \midrule
        Simple & 0.437 & 7.947 & 5.298 & 0.166 & 0.973 & 0.804 & 0.214 & 0.057 & 0.000 & -1.000 & 0.000 & \bf{0.295} & 0.340 & 0.000 \\
        ProCoT-p4g & 0.513 & 7.497 & 5.123 & 0.192 & 0.959 & 0.911 & 0.250 & 0.245 & 0.097 & -1.000 & 0.000 & 0.095 & 0.525 & 0.107 \\
        ProCoT-p4g-desc & 0.437 & 7.710 & \bf{4.756} & \bf{0.308} & 0.959 & 0.768 & 0.196 & 0.075 & 0.048 & -1.000 & 0.000 & 0.289 & \bf{0.633} & \bf{0.186} \\
        \cmidrule{1-15}
        ProCoT-rich & 0.510 & 7.530 & 5.157 & 0.211 & 0.973 & 0.821 & 0.321 & 0.226 & 0.097 & -1.500 & 0.000 & 0.211 & 0.415 & 0.161 \\
        ProCoT-rich-desc & \bf{0.603} & \bf{7.187} & 5.337 & 0.185 & \bf{1.000} & \bf{0.946} & \bf{0.357} & \bf{0.396} & \bf{0.226} & - & 0.000 & 0.111 & 0.375 & 0.125 \\
        \bottomrule
    \end{tabular}
    \caption{Performance by persuasive agent patterns in English.}
    \label{tab:model_comparison_split_init_category_en}
\end{table*}

\setlength{\tabcolsep}{6pt} 

\begin{table*}[t]
    \centering
    \small
    \tabcolsep 3pt
    \begin{tabular}{l|cccc|cccc}
        \toprule
        \multirow{2}{*}{\bf{Persuasive Strategy}} & \multicolumn{4}{c|}{\bf Japanese} & \multicolumn{4}{c}{\bf English} \\ 
        & \bf{PC-p} & \bf{PC-pd} & \bf{PC-r} & \bf{PC-rd} & \bf{PC-p} & \bf{PC-pd} & \bf{PC-r} & \bf{PC-rd} \\
        \midrule
        Emotion appeal & \bf{0.488} & \bf{0.281} & \bf{0.282} & \bf{0.247} & \bf{0.294} & 0.165 & \bf{0.254} & 0.165 \\ 
        Foot in the door & 0.221 & 0.144 & 0.241 & 0.241 & 0.232 & \bf{0.234} & 0.246 & \bf{0.233} \\ 
        Donation information & 0.131 & 0.128 & - & - & 0.107 & 0.058 & - & - \\ 
        Credibility appeal & 0.063 & 0.189 & 0.038 & 0.058 & 0.120 & 0.205 & 0.076 & 0.091 \\ 
        Logical appeal & 0.039 & 0.048 & 0.031 & 0.127 & 0.051 & 0.162 & 0.054 & 0.176 \\ 
        Detailed information & - & - & 0.099 & 0.040 & - & - & 0.096 & 0.011 \\ 
        Source-related inquiry & 0.006 & 0.151 & 0.000 & 0.168 & 0.002 & 0.001 & 0.000 & 0.131 \\ 
        Self-modeling & 0.029 & 0.023 & - & - & 0.123 & 0.044 & - & - \\ 
        Social proof & - & - & 0.054 & 0.004 & - & - & 0.063 & 0.010 \\ 
        Time pressure & - & - & 0.042 & 0.016 & - & - & 0.062 & 0.008 \\ 
        Personal story & 0.020 & 0.023 & 0.003 & 0.000 & 0.067 & 0.063 & 0.038 & 0.002 \\ 
        Feedback and thanks & - & - & 0.017 & 0.023 & - & - & 0.026 & 0.027 \\ 
        Personal-related inquiry & 0.001 & 0.004 & 0.171 & 0.000 & 0.002 & 0.003 & 0.000 & 0.000 \\ 
        Consistency & - & - & 0.001 & 0.023 & - & - & 0.005 & 0.053 \\ 
        Scarcity & - & - & 0.004 & 0.013 & - & - & 0.027 & 0.033 \\ 
        Personal relevance emphasis & - & - & 0.003 & 0.011 & - & - & 0.024 & 0.018 \\ 
        Task-related inquiry & 0.004 & 0.008 & 0.000 & 0.000 & 0.002 & 0.064 & 0.001 & 0.000 \\ 
        Message strength & - & - & 0.006 & 0.015 & - & - & 0.004 & 0.013 \\ 
        Framing & - & - & 0.001 & 0.012 & - & - & 0.008 & 0.012 \\ 
        Mutual concession & - & - & 0.000 & 0.001 & - & - & 0.000 & 0.014 \\ 
        Shared Engagement & - & - & 0.003 & 0.000 & - & - & 0.009 & 0.000 \\ 
        Ongoing trust building & - & - & 0.005 & 0.001 & - & - & 0.003 & 0.000 \\ 
        Reciprocity & - & - & 0.001 & 0.000 & - & - & 0.005 & 0.001 \\ 
        Metacognitive approach & - & - & 0.000 & 0.001 & - & - & 0.000 & 0.002 \\ 
        Continuous communication & - & - & 0.000 & 0.000 & - & - & 0.000 & 0.000 \\ 
        Manage cognitive load & - & - & 0.000 & 0.000 & - & - & 0.000 & 0.000 \\ 
        \midrule
        Entropy (w/o unused str.) & 2.167 & 2.738 & 2.871 & 2.973 & 2.647 & 2.814 & 3.202 & 3.202 \\
        Entropy (all) & 2.157 & 2.721 & 0.194 & 0.181 & 2.631 & 2.794 & 0.149 & 0.154 \\
        \bottomrule
    \end{tabular}
    \caption{Detailed distributions of strategies used by each agent. \textbf{PC-p}, \textbf{PC-pd}, \textbf{PC-r}, and \textbf{PC-rd} represent the shortened notations for \textbf{\textit{ProCoT-p4g}}, \textbf{\textit{ProCoT-p4g-desc}}, \textbf{\textit{ProCoT-rich}}, and \textbf{\textit{ProCoT-rich-desc}}, respectively.}
    \label{tab:agent_strategy_analisis_both}
\end{table*}

\setlength{\tabcolsep}{3pt} 

\begin{table*}[ht]
    \footnotesize
    \centering
    \begin{tabular}{p{1.3cm}l|ccccc|ccccc|ccccc|ccccc|ccccc}
        \toprule
        \multicolumn{2}{c|}{\multirow{3}{*}{\bf{Agent Pattern}}} & \multicolumn{25}{c}{\bf{Donation Intention (Top Row: Initial One, Second Row: Final One)}} \\
        & & \bf{1} & & & & & \bf{2} & & & & & \bf{3} & & & & & \bf{4} & & & & & \bf{5} & & & & \\
        & & \bf{1} & \bf{2} & \bf{3} & \bf{4} & \bf{5} & \bf{1} & \bf{2} & \bf{3} & \bf{4} & \bf{5} & \bf{1} & \bf{2} & \bf{3} & \bf{4} & \bf{5} & \bf{1} & \bf{2} & \bf{3} & \bf{4} & \bf{5} & \bf{1} & \bf{2} & \bf{3} & \bf{4} & \bf{5} \\
        \midrule
        \multirow{5}{0.5cm}{Japanese} & Simple & \bf{70} & 3 & 0 & 0 & 0 & \bf{53} & 3 & 0 & 0 & 0 & \bf{25} & 12 & 19 & 0 & 0 & 13 & 4 & 8 & \bf{27} & 1 & 5 & 3 & 1 & 0 & \bf{53} \\
        & ProCoT-p4g & \bf{72} & 1 & 0 & 0 & 0 & \bf{56} & 0 & 0 & 0 & 0 & \bf{45} & 3 & 8 & 0 & 0 & \bf{34} & 1 & 7 & 11 & 0 & 21 & 1 & 8 & 0 & \bf{32} \\
        & ProCoT-p4g-desc & \bf{73} & 0 & 0 & 0 & 0 & \bf{53} & 3 & 0 & 0 & 0 & \bf{46} & 5 & 5 & 0 & 0 & \bf{25} & 4 & 7 & 17 & 0 & 15 & 1 & 1 & 3 & \bf{42} \\
        & ProCoT-rich & \bf{73} & 0 & 0 & 0 & 0 & \bf{56} & 0 & 0 & 0 & 0 & \bf{50} & 1 & 5 & 0 & 0 & \bf{41} & 2 & 1 & 9 & 0 & 29 & 1 & 1 & 0 & \bf{31} \\
        & ProCoT-rich-desc & \bf{73} & 0 & 0 & 0 & 0 & \bf{53} & 3 & 0 & 0 & 0 & \bf{50} & 2 & 4 & 0 & 0 & \bf{43} & 3 & 5 & 2 & 0 & \bf{31} & 0 & 5 & 1 & 25 \\
        \midrule
        \multirow{5}{0.5cm}{English} & simple & \bf{71} & 2 & 0 & 0 & 0 & \bf{45} & 11 & 0 & 0 & 0 & 12 & 13 & \bf{31} & 0 & 0 & 3 & 2 & 13 & \bf{35} & 0 & 0 & 0 & 0 & 0 & \bf{62} \\
        & ProCoT & \bf{70} & 3 & 0 & 0 & 0 & \bf{51} & 5 & 0 & 0 & 0 & 14 & 4 & \bf{38} & 0 & 0 & 13 & 7 & 7 & \bf{26} & 0 & 6 & 0 & 3 & 0 & \bf{53} \\
        & ProCoT-desc & \bf{70} & 3 & 0 & 0 & 0 & \bf{43} & 13 & 0 & 0 & 0 & 11 & 13 & \bf{32} & 0 & 0 & 4 & 7 & 17 & \bf{25} & 0 & 3 & 1 & 4 & 0 & \bf{54} \\
        & ProCoT-rich & \bf{71} & 1 & 1 & 0 & 0 & \bf{46} & 10 & 0 & 0 & 0 & 18 & 8 & \bf{30} & 0 & 0 & 12 & 4 & 9 & \bf{28} & 0 & 6 & 1 & 3 & 0 & \bf{52} \\
        & ProCoT-rich-desc & \bf{73} & 0 & 0 & 0 & 0 & \bf{53} & 3 & 0 & 0 & 0 & 20 & 4 & \bf{32} & 0 & 0 & \bf{21} & 1 & 10 & \bf{21} & 0 & 14 & 0 & 3 & 0 & \bf{45} \\
        \bottomrule
    \end{tabular}
    \caption{Detailed distributions of shifting in donation intention.}
    \label{tab:intention-shifts}
\end{table*}

\setlength{\tabcolsep}{6pt} 

\begin{table*}[t]
    \footnotesize
    \centering
    \begin{tabular}{l rrrr}
        \toprule
        \textbf{Domain} & \textbf{Win (\%)} & \textbf{Tie (\%)} & \textbf{Lose (\%)} & \textbf{Instances} \\
        \midrule
        Economics & 80.00 & 10.00 & 10.00 & 10 \\ 
        History & 75.00 & 12.50 & 12.50 & 8 \\ 
        Sport & 67.86 & 17.86 & 14.29 & 28 \\ 
        Travel & 65.71 & 22.86 & 11.43 & 35 \\ 
        Finance & 63.83 & 21.28 & 14.89 & 47 \\ 
        Research & 63.64 & 27.27 & 9.09 & 11 \\ 
        Welfare & 63.16 & 26.32 & 10.53 & 19 \\ 
        Ethics & 62.22 & 28.89 & 8.89 & 45 \\ 
        Psychology & 60.94 & 28.12 & 10.94 & 64 \\ 
        Career & 60.00 & 20.00 & 20.00 & 85 \\ 
        Negotiation & 60.00 & 40.00 & 0.00 & 5 \\ 
        Philosophy & 59.09 & 40.91 & 0.00 & 22 \\ 
        Craftsmanship & 57.14 & 0.00 & 42.86 & 7 \\ 
        Education & 56.69 & 27.39 & 15.92 & 157 \\ 
        Marketing & 55.56 & 11.11 & 33.33 & 18 \\ 
        Innovation & 53.85 & 30.77 & 15.38 & 13 \\ 
        Ecology & 52.83 & 32.08 & 15.09 & 53 \\ 
        Media & 52.63 & 26.32 & 21.05 & 19 \\ 
        Family & 50.88 & 22.81 & 26.32 & 57 \\ 
        Business & 50.65 & 35.06 & 14.29 & 77 \\ 
        Charity & 50.00 & 25.00 & 25.00 & 32 \\ 
        Science & 50.00 & 26.67 & 23.33 & 30 \\ 
        Safety & 50.00 & 27.27 & 22.73 & 22 \\ 
        Health & 49.32 & 34.25 & 16.44 & 73 \\ 
        Lifestyle & 48.97 & 29.66 & 21.38 & 145 \\ 
        Politics & 48.28 & 41.38 & 10.34 & 29 \\ 
        Architecture & 45.45 & 54.55 & 0.00 & 11 \\ 
        Technology & 45.33 & 41.33 & 13.33 & 75 \\ 
        Debate & 44.44 & 44.44 & 11.11 & 9 \\ 
        Art & 44.00 & 34.00 & 22.00 & 50 \\ 
        Culture & 42.11 & 36.84 & 21.05 & 38 \\ 
        Leisure & 41.38 & 48.28 & 10.34 & 29 \\ 
        Literature & 41.03 & 41.03 & 17.95 & 39 \\ 
        Fashion & 38.46 & 38.46 & 23.08 & 13 \\ 
        Law & 0.00 & 50.00 & 50.00 & 4 \\ 
        \bottomrule
    \end{tabular}
    \caption{Win-Rate by domain against \textit{Simple}.}
    \label{tab:result-daily-persuasion-per-domain-simple}
\end{table*}

\begin{table*}[t]
    \footnotesize
    \centering
    \begin{tabular}{l rrrr}
        \toprule
        \textbf{Domain} & \textbf{Win (\%)} & \textbf{Tie (\%)} & \textbf{Lose (\%)} & \textbf{Instances} \\
        \midrule
        Marketing & 61.11 & 22.22 & 16.67 & 18 \\ 
        Debate & 55.56 & 22.22 & 22.22 & 9 \\ 
        History & 50.00 & 25.00 & 25.00 & 8 \\ 
        Finance & 44.68 & 31.91 & 23.40 & 47 \\ 
        Travel & 42.86 & 37.14 & 20.00 & 35 \\ 
        Sport & 42.86 & 35.71 & 21.43 & 28 \\ 
        Welfare & 42.11 & 42.11 & 15.79 & 19 \\ 
        Charity & 40.62 & 34.38 & 25.00 & 32 \\ 
        Technology & 40.00 & 41.33 & 18.67 & 75 \\ 
        Economics & 40.00 & 30.00 & 30.00 & 10 \\ 
        Negotiation & 40.00 & 60.00 & 0.00 & 5 \\ 
        Psychology & 39.06 & 46.88 & 14.06 & 64 \\ 
        Innovation & 38.46 & 46.15 & 15.38 & 13 \\ 
        Ethics & 37.78 & 26.67 & 35.56 & 45 \\ 
        Culture & 36.84 & 42.11 & 21.05 & 38 \\ 
        Media & 36.84 & 21.05 & 42.11 & 19 \\ 
        Literature & 35.90 & 30.77 & 33.33 & 39 \\ 
        Politics & 34.48 & 51.72 & 13.79 & 29 \\ 
        Leisure & 34.48 & 20.69 & 44.83 & 29 \\ 
        Health & 34.25 & 49.32 & 16.44 & 73 \\ 
        Career & 34.12 & 38.82 & 27.06 & 85 \\ 
        Family & 33.33 & 38.60 & 28.07 & 57 \\ 
        Lifestyle & 32.41 & 40.69 & 26.90 & 145 \\ 
        Education & 31.85 & 38.22 & 29.94 & 157 \\ 
        Fashion & 30.77 & 23.08 & 46.15 & 13 \\ 
        Business & 28.57 & 37.66 & 33.77 & 77 \\ 
        Craftsmanship & 28.57 & 14.29 & 57.14 & 7 \\ 
        Safety & 27.27 & 36.36 & 36.36 & 22 \\ 
        Science & 26.67 & 46.67 & 26.67 & 30 \\ 
        Ecology & 26.42 & 37.74 & 35.85 & 53 \\ 
        Philosophy & 22.73 & 50.00 & 27.27 & 22 \\ 
        Art & 22.00 & 42.00 & 36.00 & 50 \\ 
        Research & 18.18 & 63.64 & 18.18 & 11 \\ 
        Architecture & 18.18 & 45.45 & 36.36 & 11 \\ 
        Law & 0.00 & 25.00 & 75.00 & 4 \\  
        \bottomrule
    \end{tabular}
    \caption{Win-Rate by domain against \textit{ProCoT-p4g}.}
    \label{tab:result-daily-persuasion-per-domain-p4g}
\end{table*}
\begin{table*}[t]
    \footnotesize
    \centering
    \begin{tabular}{l rrrr}
        \toprule
        \textbf{Turn} & \textbf{Win (\%)} & \textbf{Tie (\%)} & \textbf{Lose (\%)} & \textbf{Instances} \\
        \midrule
        1 & 32.18 & 34.48 & 33.33 & 174 \\ 
        2 & 68.47 & 26.11 & 5.42 & 203 \\ 
        3 & 52.38 & 33.81 & 13.81 & 210 \\ 
        4 & 61.58 & 27.09 & 11.33 & 203 \\ 
        5 & 54.09 & 30.82 & 15.09 & 159 \\ 
        6 & 53.33 & 35.56 & 11.11 & 45 \\ 
        7 & 60.00 & 0.00 & 40.00 & 5 \\ 
        8 & 100.00 & 0.00 & 0.00 & 1 \\ 
        \bottomrule
    \end{tabular}
    \caption{Win-Rate by target turns against \textit{Simple}.}
    \label{tab:result-daily-persuasion-per-history-turns-simple}
\end{table*}

\begin{table*}[t]
    \footnotesize
    \centering
    \begin{tabular}{l rrrr}
        \toprule
        \textbf{Turns} & \textbf{Win (\%)} & \textbf{Tie (\%)} & \textbf{Lose (\%)} & \textbf{Instances} \\
        \midrule
        1 & 24.14 & 28.16 & 47.70 & 174 \\ 
        2 & 40.39 & 37.93 & 21.67 & 203 \\ 
        3 & 35.24 & 49.05 & 15.71 & 210 \\ 
        4 & 36.45 & 41.87 & 21.67 & 203 \\ 
        5 & 37.11 & 37.11 & 25.79 & 159 \\ 
        6 & 37.78 & 37.78 & 24.44 & 45 \\ 
        7 & 40.00 & 40.00 & 20.00 & 5 \\ 
        8 & 100.00 & 0.00 & 0.00 & 1 \\ 
        \bottomrule
    \end{tabular}
    \caption{Win-Rate by target turns against \textit{ProCoT-p4g}}
    \label{tab:result-daily-persuasion-per-history-turns-p4g}
\end{table*}

\begin{table*}[ht]
    \centering
    \small
    \begin{tabular}{ll}
        \toprule
        \textbf{Background} & 
        \begin{tabular}{p{11cm}}
         Nina's birthday is coming up, and her friends want to throw her a surprise party at a local bowling alley. However, Nina's best friend, Chloe, believes that a karaoke night would be a better fit for Nina's personality.  \\
        \end{tabular}\\
        \midrule 
        \textbf{Goal} & 
        \begin{tabular}{p{11cm}}
            Persuade Nina's friends to organize a karaoke night\\
        \end{tabular}\\
        \midrule
        \multirow{3}{*}{\textbf{ProCoT-rich-desc (Uni-X)}} & 
        \begin{tabular}{p{11cm}}
            \textbf{Strategy:} A-2 Task-related inquiry  \\ 
        \end{tabular} \\
         & \begin{tabular}{p{11cm}}
          \textbf{Utterance:} What do you all think Nina would enjoy most for her birthday—something fun and interactive like karaoke or something more casual like bowling? \\ 
         \end{tabular}\\
        \midrule
        \multirow{3}{*}{\textbf{ProCoT-p4g (Uni-Y)}} & 
        \begin{tabular}{p{11cm}}
             \textbf{Strategy:} Emotion appeal \\
        \end{tabular} \\
         & \begin{tabular}{p{11cm}}
          Nina loves singing and expressing herself, so a karaoke night would make her feel truly celebrated and special. \\
         \end{tabular}\\
        \midrule
        \multirow{2}{*}{\textbf{Judge}} & 
            \begin{tabular}{p{11cm}}
            \textbf{Experience:} Given the single utterance in each universe, Uni-Y already supplies an argument and emotional appeal toward karaoke, while Uni-X only sets up a question. Thus Uni-Y is measurably more persuasive and aligned with the goal.\\
            \end{tabular}\\
         & 
         \begin{tabular}{p{11cm}}
              \textbf{Result:} Uni-Y  \\ 
         \end{tabular} \\
        \bottomrule
    \end{tabular}
    \caption{Example of losing instance at first turn against \textit{ProCoT-p4g}.}
    \label{tab:example-1turn-lose}
\end{table*}

\begin{table*}[ht]
    \centering
    \small
    \begin{tabular}{ll}
        \toprule
        \textbf{Background} & 
        \begin{tabular}{p{11cm}}
         Tom, a marketing manager, is considering investing in big data analysis to improve his company's marketing strategies. His colleague, Jane, is skeptical about the effectiveness of big data analysis and believes that traditional marketing methods are sufficient. Tom wants to persuade Jane of the benefits of big data analysis.  \\
        \end{tabular}\\
        \midrule 
        \textbf{Goal} & 
        \begin{tabular}{p{11cm}}
            Persuade Jane to support investing in big data analysis\\
        \end{tabular}\\
        \midrule
        \multirow{3}{*}{\textbf{ProCoT-rich-desc (Uni-X)}} & 
        \begin{tabular}{p{11cm}}
            \textbf{Strategy:} A-2 Task-related inquiry  \\ 
        \end{tabular} \\
         & \begin{tabular}{p{11cm}}
          \textbf{Utterance:} What do you think are the biggest strengths of traditional marketing methods, and what concerns you most about using big data analysis? \\ 
         \end{tabular}\\
        \midrule
        \multirow{4}{*}{\textbf{ProCoT-p4g (Uni-Y)}} & 
        \begin{tabular}{p{11cm}}
             \textbf{Strategy:} Logical appeal \\
        \end{tabular} \\
         & \begin{tabular}{p{11cm}}
          Big data analysis allows us to identify precise customer trends and preferences, which traditional methods often miss, leading to more effective and targeted marketing strategies. \\
         \end{tabular}\\
        \midrule
        \multirow{2}{*}{\textbf{Judge}} & 
            \begin{tabular}{p{11cm}}
            \textbf{Experience:} Although Uni-Y supplies immediate argumentative content, Uni-X sets up a more dialogic, audience-centred approach that is generally regarded as more effective in persuasion, especially at the opening of a conversation. Therefore, Uni-X’s performance is slightly better.\\
            \end{tabular}\\
         & 
         \begin{tabular}{p{11cm}}
              \textbf{Result:} Uni-X  \\ 
         \end{tabular} \\
        \bottomrule
    \end{tabular}
    \caption{Example of winning instance at first turn against \textit{ProCoT-p4g}.}
    \label{tab:example-1turn-win}
\end{table*}
\end{document}